\documentclass[preprint,1p]{elsarticle}

\journal{Expert Systems With Applications}

\usepackage[english]{babel}
\usepackage[utf8]{inputenc}  
\usepackage{epsfig}           
\usepackage{subfigure}
\usepackage{graphicx}
\usepackage{setspace}
\usepackage{amsmath}
\usepackage{amsfonts}
\usepackage{amssymb}
\usepackage{calligra}
\usepackage{eucal}
\usepackage{multirow}
\usepackage{geometry}
\usepackage{graphicx}

\usepackage{vmargin} 
\usepackage{lscape}
 
\usepackage{pifont}

\usepackage{color}
\newcommand{\todo}[1]{\textcolor{black}{#1}}

\def\x{\times}
\def\pt{p_\textrm{t}}
\def\at{\alpha_\textrm{t}}

\def\CE{\textrm{CE}}
\def\FL{\textrm{FL}}

\newcommand{\eqnnm}[2]{\begin{equation}\label{eq:#1}#2\end{equation}\ignorespaces}

\setmarginsrb {30mm} 
{20mm} 
{25mm} 
{35mm} 
{2ex}  
{5ex}  
{0pt}  
{0mm}  

\begin {document}

\begin{frontmatter}

\title{A Comprehensive Comparison of End-to-End Approaches for Handwritten Digit String Recognition}
\vspace{-3mm}
\author[pucpr]{Andre~G.~Hochuli\corref{corr}}
\cortext[corr]{Corresponding author}
\ead{aghochuli@ppgia.pucpr.br}
\author[pucpr]{Alceu~S.~Britto Jr}
\ead{alceu@ppgia.pucpr.br}
\author[orand]{David~A.~Saji}
\ead{david.saji@ing.uchile.cl}
\author[orand]{José~M.~Saavedra}
\ead{jose.saavedra@orand.cl}
\author[ets]{\\Robert~Sabourin}
\ead{robert.sabourin@etsmtl.ca }
\author[ufpr]{Luiz~S.~Oliveira}
\ead{luiz.oliveira@ufpr.br}

\address[pucpr]{Pontifical Catholic University of Parana (PUCPR), Curitiba, Brazil \\
		R. Imaculada Concei\c c\~ao, 1155, Curitiba, PR, Brazil - 80215-901\\[0.7ex]}

\address[orand]{Computer Vision Research Group, ORAND S.A \\
		Estado 360, Of 702, Santiago, Chile\\[0.7ex]}

\address[ets]{École de Technologie Supérieure (ÉTS) \\
		1100 Notre Dame West, Montreal, Quebec, Canada\\[0.7ex]}

\address[ufpr]{Federal University of Parana (UFPR), Curitiba, Brazil \\
		Rua Cel. Francisco H. dos Santos, 100, PR, Brazil - 81531-990\\[-5ex] }

\begin{abstract}

Over the last decades, most approaches \todo{proposed} for handwritten digit string recognition (HDSR) \textcolor{black}{have} \textcolor{black}{resorted} to digit segmentation, which is dominated by heuristics, \textcolor{black}{thereby} imposing substantial constraints on the final performance. Few of them \textcolor{black}{have been} based on segmentation-free strategies where each pixel column has a potential cut location. Recently, segmentation-free strategies \textcolor{black}{has added} another perspective to the problem, \textcolor{black}{leading} to promising results. However, these strategies still show some limitations when dealing with a large number of touching digits. \textcolor{black}{To bridge the resulting gap}, in this paper, we hypothesize that a string of digits can be approached as a sequence of objects. \textcolor{black}{We thus} evaluate different end-to-end approaches to solve the HDSR problem, particularly in two verticals: those based on object-detection (e.g., Yolo and RetinaNet) and those based on sequence-to-sequence representation (CRNN).

The main contribution \textcolor{black}{of this work lies in its provision} of a comprehensive comparison with a critical analysis of the above mentioned strategies on five benchmarks commonly used to assess HDSR, including the challenging  Touching Pair dataset, NIST SD19, and two real-world datasets (CAR and CVL) proposed for the ICFHR 2014 competition on HDSR. Our results show that the Yolo model compares favorably \textcolor{black}{against} segmentation-free models with the advantage of having a shorter pipeline that minimizes the presence of heuristics-based models. It achieved a 97\%, 96\%, and 84\% recognition rate on the NIST-SD19, CAR, and CVL datasets, respectively.

\end{abstract}


\end{frontmatter}

\section{Introduction}
	
\textcolor{black}{Research }in handwritten digit string recognition (HDSR) has \textcolor{black}{picked up over} the past few decades. \textcolor{black}{Most works covering the subject share a common strategy, which involves segmenting a string into isolated digits and then applying} a classifier capable of recognizing 10 classes (0...9). However, a straightforward solution becomes unfeasible in the presence of noise, broken digits, and in the worst case, touching digits. The impacts of the first two cases are reduced \textcolor{black}{when} some heuristic-based pre-processing modules \textcolor{black}{are applied}. The challenge, \textcolor{black}{however} remains over touching digits.
    
\begin{figure}[htbp]
	\begin{center}
		\mbox{
			\subfigure[]{\scalebox{0.65}{\epsffile{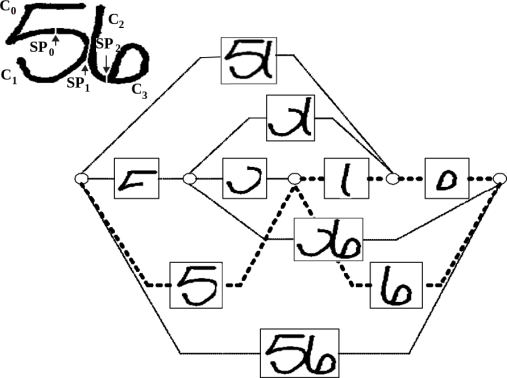}}} \quad
			\subfigure[]{\scalebox{0.25}{\epsffile{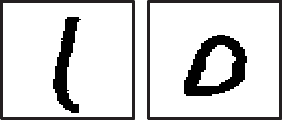}}}
		}
		\caption{(a) Segmentation paths for the string ``56'' and (b) Images that can be easily confused with digits ``0'' and ``1'' (extracted from \cite{Vellasques2008}). }
		\label{seggraph:fig}
	\end{center}
\end{figure}

To handle the presence of touching digits, algorithms based on contour and profile information over segment the numerical string, generating components that may represent a digit or part of it. After each resulting component \textcolor{black}{is classified}, a fusion method determines the best combination among \textcolor{black}{many} hypotheses. The rationale behind over-segmentation is to maximize the chances of producing the correct segmentation, \textcolor{black}{even at a high post-processing computational cost}. This strategy is illustrated in Figure \ref{seggraph:fig}. Readers interested in different global and local approaches may refer to \cite{Casey96} and \cite{Ribas2013}. These two works survey the state-of-the-art up to 2012, \textcolor{black}{while} the approaches proposed by \cite{Gattal2015} and \cite{Gattal2017} were the last attempts using the segmentation-based approach.

The alternative approaches resort to segmentation-free based methods (\cite{Choi99,Procter98,Alceu2001,Ciresan2012,Hochuli2018b}) in which the string is recognized without the need for its a priori segmentation into isolated digits. \textcolor{black}{This approach only recently started gaining attention among the research community, prodded by advances in machine learning thanks to deep learning techniques}. While over-segmentation based methods demand \textcolor{black}{certain} specific strategies to generate segmentation cuts, a robust isolated digit recognizer, \textcolor{black}{as well as} a strategy for searching the best path among the generated segmentation hypothesis, the segmentation-free demands a significant amount of training data. \textcolor{black}{Both strategies are caracterized by} common complexes pipelines surrounded by handcrafted features, heuristic modules, and fusion rules to assembly task-specific classifiers. The \textcolor{black}{need} for an end-to-end approach is \textcolor{black}{therefore} evident.
         
\textcolor{black}{Contrary to} the handwritten digit string recognition, the object recognition \textcolor{black}{field is evolving} very rapidly. \textcolor{black}{Each} year, new algorithms \textcolor{black}{surface and outperform} the previous ones. Consequently, there \textcolor{black}{presently} are a plethora of \textcolor{black}{ready-to-use} pre-trained deep learning end-to-end models \textcolor{black}{available} (\cite{ YOLO2016,  YOLO2017, RCNN, FastRCNN, FasterRCNN, RetinaNet2017}). In the same vein, sequence-to-sequence based models (\cite{Voigtlaender2016,Shi2017-CRNN, Dutta2018}) have produced end-to-end solutions for temporal series, handwritten text, and text scene recognition. Besides high performance, these approaches \textcolor{black}{contribute significantly} by providing a reduced number of handcrafted features and heuristics methods, producing a straightforward pipeline as compared to related state-of-the-art works.


In discussing end-to-end approaches, one aspect that is very often highlighted in the literature is the importance of context. Several recent computer vision approaches have demonstrated that the use of context improves recognition performance (\cite{divvala2009empirical}). In the case of digit string recognition, contextual information is more limited, but it \textcolor{black}{nonetheless} plays a vital role, as demonstrated in \cite{Oliveira02b}. 

In this paper, we argue that a string of digits is a sequence of objects. Therefore, we restricted our scope to the following neural network-based approaches: a) Yolo (\cite{ YOLO2016,  YOLO2017}), b) RetinaNet (\cite{RetinaNet2017}) which \textcolor{black}{is a} state-of-the-art architectures for object detection/recognition, and c) CRNN (\cite{Shi2017-CRNN}), a sequence-to-sequence model composed of a convolutional network combined with a long-short term memory (LSTM) (\cite{Schuster1997}). To complete our analysis, we also consider two approaches based on dynamic selection (\cite{Hochuli2018} and \cite{Aly2019}). To deploy end-to-end approaches for this problem, we generate a large dataset of strings mimicking real datasets, \textcolor{black}{which provides} contextual information for training. Even though \cite{Zhan2017} applied CRNN for courtesy amount recognition on bank checks, we provide an in-depth analysis of this model \textcolor{black}{based on} different challenging benchmarks, \textcolor{black}{and compare} it with other end-to-end approaches, \textcolor{black}{such as those that are object detection-based}.

The main contributions of this work \textcolor{black}{lies in its provision of} a comprehensive comparison, \textcolor{black}{along with a} critical analysis of the end-to-end object recognition strategies, sequence-to-sequence approaches used for handwritten words, and the recently published specific segmentation-free HDSR methods. 
Our extensive experimental protocol include experiments on the following benchmarks: i) Touching Pair (TP) dataset (\cite{Ribas2013}), which contains 79,464 touching digits and has been used a benchmark for both heuristic-based and segmentation-free algorithms; ii) 570,000 images of strings composed of 2-, 3-, and 4-touching digits; iii) NIST SD19, which is composed of 11,585 real-world numerical strings, ranging from 2 to 6 digits, and iv) ICFHR 2014 competition (\cite{Diem2014}), which contains real courtesy amount of bank checks and a \textcolor{black}{significant} variability of handwritten styles.

\textcolor{black}{Our} experimental analysis shows the limits of the proposed strategies for the HDSR. End-to-end approaches, especially in the Yolo model, compare favorably \textcolor{black}{against} the segmentation-free methods in \cite{Hochuli2018, Hochuli2018b} with the clear advantage of having a shorter pipeline that minimizes the presence of heuristic-based modules, such as \textcolor{black}{those} pre-processing. On the other hand,  bottlenecks \textcolor{black}{associated with} the laborious task of annotation of ground-truths when synthetic data are not applicable and the lack of lexicon for digit strings is a matter of discussion.
    
This paper is organized as follows: Section \ref{related_work:sec} \textcolor{black}{examines} related works. The problem statement is presented in Section \ref{problemst:sec}. A detailed review of architectures is given in Section \ref{hdsr_app:sec}. In Section \ref{sec:Experiments}, we \textcolor{black}{tackle} the approaches using the aforementioned datasets. Finally, Section \ref{conclusions:sec} concludes this work.

\section{Related Works}\label{related_work:sec}


To avoid the burden of over-segmentation, some authors have devoted efforts towards segmentation-free approaches. To the best of our knowledge, the first attempt in this direction was in the Space Displacement Neural Network (SDNN) introduced by  \cite{Matan92}. This strategy produces a series of output vectors used by a post-processor to \textcolor{black}{extract} the best possible label sequence from the vector sequence. As stated by \cite{LeCun98}, SDNN is an attractive technique but has not \textcolor{black}{managed to yield} better results than heuristic over-segmentation methods.

\textcolor{black}{The} Hidden Markov Model (HMM), initially developed in the field of speech recognition, \textcolor{black}{has been} used to build segmentation-free methods for handwriting recognition. \cite{Elms98} first applied HMM to word recognition and then adapted their work to classify handwritten digit strings of unknown length (\cite{Procter98}). \cite{Alceu2001} revisited these two studies and proposed a two-stage segmentation-free method using features extracted from lines and columns that are processed by a set of HMMs. This framework achieved an average recognition rate of 91.0\% in NIST-SD19.
	
\cite{Choi99} designed a framework based on 100 neural networks to avoid the segmentation of touching pairs. Their approach achieves 95.3\% of the recognition rate of touching pairs extracted from NIST-SD19 (\cite{NISTSD192016}). A decade later, \cite{Ciresan2008} took advantage of Convolutional Neural Networks by training two CNNs, one for isolated digits and one for touching pairs. The authors combined these two networks to recognize 3-digit strings of \textcolor{black}{the} NIST database achieving a 93.4\% recognition rate. At that time, strings with three digits connected were not considered.

Another decade later, advances in the field of machine learning, especially with the popularization and better understanding of deep learning techniques (\cite{bengio2013,Gu2017}), \textcolor{black}{lead to} advances in different areas of handwriting recognition, such as digit recognition (\cite{Sarkhel2016,Sabour2017}), character recognition (\cite{Xiao2017, Laroca2018, Laroca2019}), word recognition (\cite{Roy2016,Tamen2017,Wua2017}), script identification (\cite{Ziyong2017}), and signature verification (\cite{Hafemann2017}). \textcolor{black}{Leveraging this evolution}, \cite{Hochuli2018} introduced a segmentation-free approach capable of recognizing digit strings of any size. In their work, the authors combined four CNNs into a Dynamic Selection (DS) scheme (\cite{BRITTO2014, CRUZ2018}). The first CNN works as a high-level classifier that determines the size of components, while the other three operate \textcolor{black}{at} a low-level by classifying 1-digit, 2-digit, and 3-digit components, respectively. This approach achieved the state-of-the-art for NIST-SD19 and Touching Pairs (\cite{Ribas2013}) datasets, surpassing segmentation-based and segmentation-free methods.
    
Despite \textcolor{black}{this} good performance, this approach has \textcolor{black}{certain limitations}. First, it is based on a hierarchical framework composed of heuristic-based pre-processing and four classifiers, \textcolor{black}{which leads to various error sources}. Second, the strategy recognizes strings of any size but limited to 3-digit touching. To mitigate some of these problems, \cite{Hochuli2018b} reduced the number of classifiers by introducing a single classifier ($\mathcal{C}_{1110}$) capable \textcolor{black}{of classifying} 1110 classes ($0 \ldots 9, 00 \ldots 99, \mbox{ and } 000 \ldots 999$). Although these approaches achieve high recognition rates, \textcolor{black}{they are still carried by} complex pipelines, \textcolor{black}{and are} surrounded by heuristic processes, pre-processing modules, and fusion strategies. 

Recently, sequence-to-sequence architectures have been successfully applied to the tasks of handwritten text recognition and scene text recognition (\cite{Voigtlaender2016,Shi2017-CRNN,Dutta2018}). Those solutions combine a Convolutional Neural Network (CNN) and a Recurrent Neural Network (RNN) to produce a sequence of probabilities interpreted by a transcription layer. This pipeline produces an end-to-end trainable model \textcolor{black}{which achieves} state-of-art performance of \textcolor{black}{handwritten text recognition}. However, it relies on a \textcolor{black}{specific} lexicon to mitigate confusions. 

\textcolor{black}{In} object recognition, the main goal \textcolor{black}{is to detect and recognize} a set of predefined classes of objects in a given input image.  Until the last decade, a classical approach \textcolor{black}{used to be} based on a sliding window and its variants (\cite{Lampert2008,Felzenszwalb2008, Felzenszwalb2010}). This approach uses a classifier trained with handcrafted features at several spatial locations of the image. A \textcolor{black}{limitation} is the high number of windows \textcolor{black}{needed} to search over multiple scales and aspect ratios. Moreover, in this exhaustive search strategy, the computational cost increases very \textcolor{black}{rapidity}.
    
A breakthrough occurred due to the arising of large-scale datasets (\cite{ILSVRC15, MSCOCO}), the popularization of GPUs and the popularization of deep networks in the ILSVRC 2012 (\cite{ILSVRC15}). At that time, this field had recovered the attention of the research community, and several deep learning-based methods were proposed \textcolor{black}{to improve} the state-of-art (\cite{Han2018}). 
     
One of the first successful approaches \textcolor{black}{in this regard consisted} of the Region-based Convolutional Network (R-CNN) proposed by \cite{RCNN}. This architecture \textcolor{black}{begins by extracting} region proposals from the image space using the selective search algorithm (\cite{SelectiveSearch}). Then, each region is warped to a fixed size, and a CNN extracts features. Finally, an SVM classifier determines a class, and a bounding-box regressor refines the locations. \textcolor{black}{The main drawback of this strategy is that it requires the extraction of} features of each warped region proposal, which is computationally expensive. 
    
To overcome this obstacle, SPPnet (\cite{SPPnet}) and Fast-RCNN (\cite{FastRCNN}) have been proposed. These models predict region proposals direct over feature maps. A spatial pooling layer is introduced to produce fixed-length representations (wrapping at feature level). Although these strategies speed up the \textcolor{black}{entire} process, they still rely on a handcrafted region proposal method. To \textcolor{black}{overcome this limitation}, \cite{MaskRCNN} introduced a region proposal network (RPN), which implicit produces candidate locations. \textcolor{black}{With this approach}, the features produced by the last convolutional layer are used on both (a) region proposal and (b) region classification \textcolor{black}{tasks}. 
    
\textcolor{black}{Despite their advatages}, the above approaches must still handle a two-stage pipeline \textcolor{black}{whenever} a region proposal strategy \textcolor{black}{is needed}, regardless of \textcolor{black}{whether or not this need} is implicit. A more ingenious alternative was proposed by \cite{YOLO2016} with the Yolo architecture, in which the authors proposed a regression-based approach that encapsulates all stages into a single network. With a single forward pass, the network provides bounding box locations and class probabilities. An essential aspect of Yolo is that it can encode the context and appearance from the neighborhood of objects, which is an important feature for implicit digit segmentation. A year \textcolor{black}{later}, the RetinaNet (\cite{RetinaNet2017}) was proposed and add a Feature Pyramid Network (FPN) to produce multi-scale features. \textcolor{black}{Its novelty lay in its introduction of} an improved loss function \textcolor{black}{known as} \textit{focal loss} to deal with class imbalance among background and foreground samples, which stifles the learning process as most image locations contain no objects. Although the RetinaNet achieves the state-of-art in object detection benchmarks, Yolo provides a good tradeoff between speed and accuracy.

\section{Problem Statement}\label{problemst:sec}	 

As stated \textcolor{black}{earlier}, traditional approaches address the problem by grouping foreground pixels into connected components, and then classifying them. \textcolor{black}{The main problem with in scenario is that when a group of pixels is extracted from an image, only a local view of the problem is obtained, with a lot of contextual information eliminated}. Without this valuable information, the algorithms suffer from the presence of noise and touching digits.

An end-to-end approach addresses this problem \textcolor{black}{holistically}. Deep learning models can learn the interaction between digits in the context of an image, which contains noise, touching, overlapping, and broken digits. Therefore, end-to-end approaches usually have short pipelines: the object detector $\mathcal{D}$ receives as input an image $I$ containing $n$ digits (objects) and produces as output the location (bounding boxes) and the digit classes $[0, \ldots, 9]$ associated with an estimation of the posterior probability. Considering that the input image $I$ may contain $n$ connected components, the most probable interpretation of the written amount $M$ is given by Equation \ref{eq:prob2}. It is worth \textcolor{black}{noting} that the CRNN approach does not provide bounding box locations because it does not implement bounding box regressors. However, the digit's location may be estimated by the receptive fields of the feature sequence (Figure \ref{crnn_pipe:fig}b).

\begin{equation}
\label{eq:prob2}
P(M|I) = \prod_{i=1}^n P(\omega_j|x_i)
\end{equation}

\noindent where $\omega_i = \{0 \ldots 9\}$ and $x_i$ stands for the digits candidates.

\section{End-to-End Strategies for HDSR}\label{hdsr_app:sec}

In this section, we present all the approaches evaluated in our work. Section \ref{ds-approach:sec} describes the dynamic selection approaches proposed by \cite{Hochuli2018} and \cite{Aly2019}, which \textcolor{black}{represented} a breakthrough in the HDSR field \textcolor{black}{as they introduced} a set of classifiers to produce a segmentation-free solution for the HDSR field. Section \ref{object-detection-models:sec} describes the object detection approaches (Yolo and RetinaNet), while Section \ref{crnn:sec}  describes the sequence-to-sequence framework (CRNN). The training protocol used for all models is presented in Section \ref{YoLo_trainning:sec}.

To \textcolor{black}{ensure} a fair evaluation, we used the source code provided by the authors whenever they were available. The repositories for the approaches reported in \cite{Hochuli2018}, \cite{YOLO2016} and \cite{RetinaNet2017} are available in \footnote{https://github.com/andrehochuli/digitstringrecognition},\footnote{https://pjreddie.com/darknet/yolov2/}, and \footnote{https://github.com/facebookresearch/Detectron}, respectively. In the case of the CRNN, the original code\footnote{https://github.com/bgshih/crnn} \textcolor{black}{was} outdated, and therefore, we used a more recent version\footnote{https://github.com/yalecyu/crnn.caffe}. \cite{Aly2019} did not share their source code, \textcolor{black}{and as a result}, in this paper, we replicate the results reported by the authors.

\subsection{Dynamic Selection Approaches}\label{ds-approach:sec}

\begin{figure}[!ht]
	\centering
	\subfigure[] {\epsfig {file=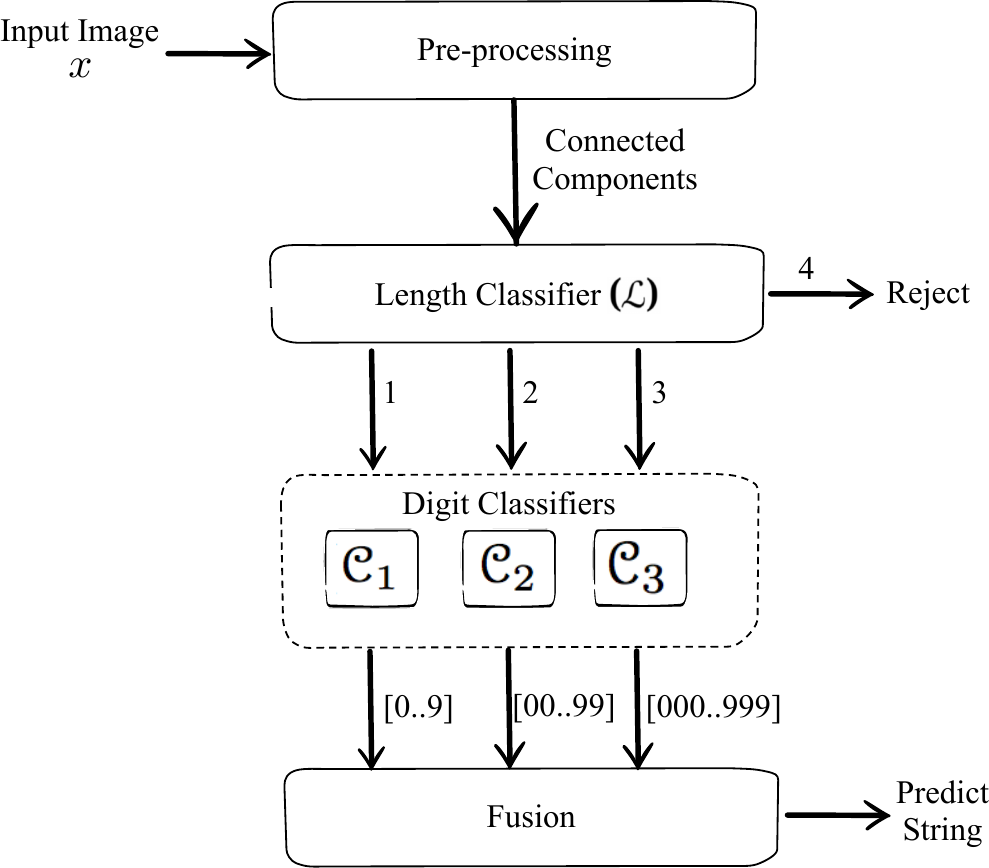, width=.5\textwidth}}
	\hspace{5mm}
	\subfigure[] {\epsfig {file=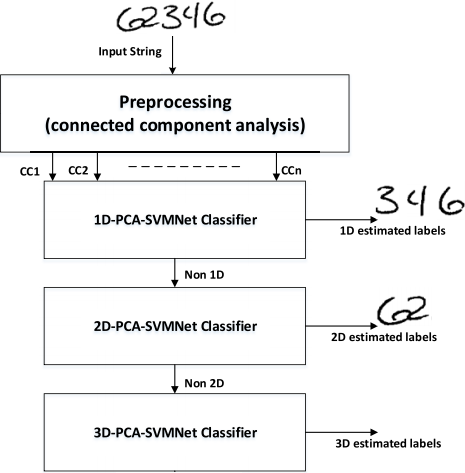, width=.45\textwidth}}
	\caption{Dynamic Selection approaches proposed by (a) \cite{Hochuli2018} and (b) \cite{Aly2019}.}
	\label{hochuli-overview:fig}	
\end{figure}

The dynamic selection framework proposed by \cite{Hochuli2018} is depicted in Figure \ref{hochuli-overview:fig}a. \textcolor{black}{Here}, a digit string $x$ is first classified by the Length classifier ($\mathcal{L}$), which will assign a probability of having 1, 2, 3, or 4 touching digits. The digit classification module comprises three classifiers ($\mathcal{C}_{1}$, $\mathcal{C}_{2}$, $\mathcal{C}_{3}$) designed to discriminate 10 [$0 \ldots 9$], 100 [$00 \ldots 99$], and 1000 [$000 \ldots 999$] classes. The classifiers that will be used for a given image depend on the output of the Length Classifier. \textcolor{black}{In accordance with} a fusion rule, more than one digit classifier may be invoked to mitigate any possible confusion.

The fusion rule used in this case considers the Top-2 outputs of $\mathcal{L}$. Let $\mathcal{L}^i(x) = p^i(x)$ be the probability of the input pattern, \textcolor{black}{and let} $x$ be composed of $i, (i = 1,2,3,4)$ digits. Let $\mathcal{C}_{1}(x) = \max\limits_{0 \leq i \leq 9} p^i(x)$, $\mathcal{C}_{2}(x) = \max\limits_{0 \leq i \leq 99} p^i(x)$, and $\mathcal{C}_{3}(x) = \max\limits_{0 \leq i \leq 999} p^i(x)$ be the probability produced by 10-class, 100-class, and 1000-class classifiers, respectively, for the input pattern $x$. Let Top1($\mathcal{C}$) and Top2($\mathcal{C}$) be the functions that return the classes with first and second highest scores of a given classifier $\mathcal{C}$, respectively. Then, $x$ is assigned to the class $\omega \in [0...1110]$, according to Equation \ref{eq:prob1},

\begin{equation}
P(\omega|x)  \left \{ \begin{array}{ll}
\mbox{if } \mathcal{L}_(x) < T, & \max(\mathcal{C}_{Top1(\mathcal{L})}(x), \mathcal{C}_{Top2(\mathcal{L})}(x)) \\
\mbox{otherwise, }    & \mathcal{C}_{Top1(\mathcal{L})}(x)\\
\end{array}
\right.
\label{eq:prob1}
\end{equation}


\noindent where $T$ is a threshold defined empirically on the validation set. 

The authors justify dealing with 1, 2, and 3 touching digits because most of the touching  occurs between two digits and sometimes between three digits (\cite{Wang00}). Strings composed of more than three touching digits are rare in real problems, \textcolor{black}{and where one occurs, it is rejected by $\mathcal{L}$.}

An alternative approach, \textcolor{black}{depicted in Figure \ref{hochuli-overview:fig}b, was proposed by \cite{Aly2019}}. In this case, the length classifier and the fusion rule were eliminated by a cascade architecture of PCA-SVMNet classifiers, which is a combination of PCA-Convolutional layers used to extract features and a linear multi-class SVM to predict classes. An extra class was introduced on each classifier as rejection, i.e., for the isolated digit classifier (10[0...9]), the class `11` contains samples of touching digits ([00...999]). The number of classes of each SVM classifier increases according to the level on the cascade.

\subsection{Object Detection Approaches}\label{object-detection-models:sec}

\begin{figure}[h!t]
	\centering
	\epsfig {file=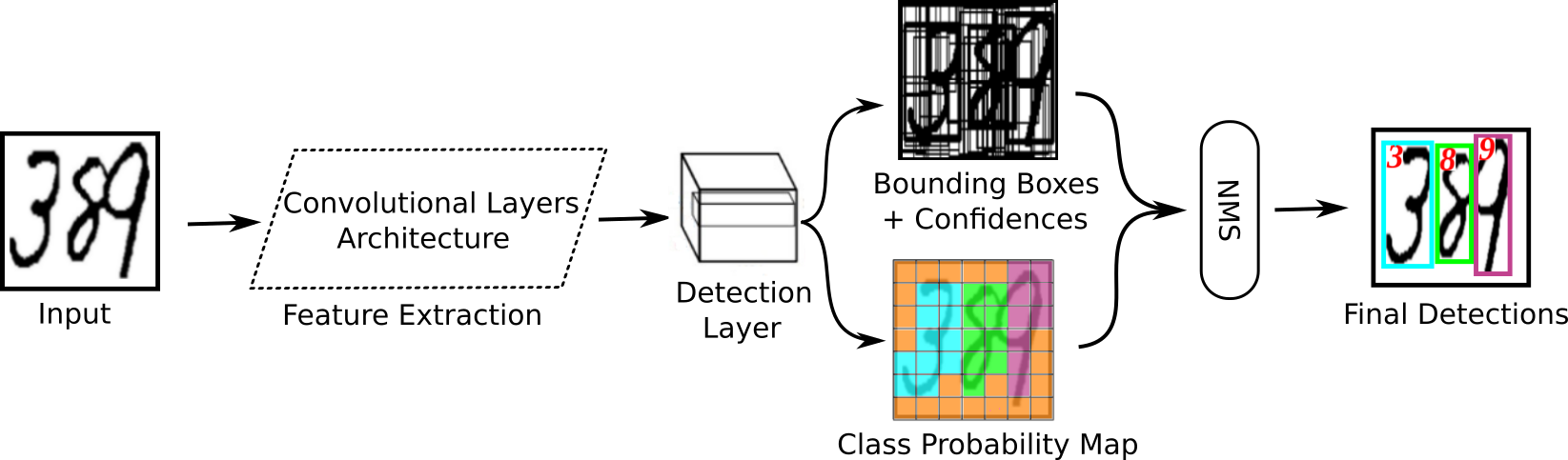, width=.95\textwidth}
	\caption{The Yolo framework divides the image into a grid and for each cell predicts bounding boxes and classes.}
	\label{YoLoframework:fig}
\end{figure}

Yolo (\cite{YOLO2016}) is a general-purpose object detection framework \textcolor{black}{that can be trained in an end-to-end fashion}. Using a single network and looking at the entire image, it can predict bounding boxes and classes with a single forward pass instead of applying the model at every location \textcolor{black}{as in the case with} traditional sliding window or region purpose-based methods (\cite{FastRCNN, FasterRCNN}). The framework is illustrated in Figure \ref{YoLoframework:fig}.

First, the convolutional layers (see Section \ref{YoLo_layers:sec}) extract features from the entire image, and then the detection layer divides the image into a grid. Next, each grid cell predicts \textcolor{black}{the coordinates of bounding boxes}, and the confidence of \textcolor{black}{each} box encloses an object. Handpicked anchor boxes are \textcolor{black}{preliminary} defined to help the network learn how to predict the right bounding boxes. Moreover, it provides class probabilities for \textcolor{black}{the cells belonging} to a \textcolor{black}{given} object. Finally, to mitigate confusion among overlapped boxes, the Non-Maximum Suppression (NMS) algorithm is used.    

The input resolution of the Darknet reported in \cite{ YOLO2016} is $416 \times 416$. However, given that strings of digits are usually wider than higher, we used an initial input size of $128 \times 256$ (height $\times$ width) to train the model. It is worth mentioning, though, that this architecture does not set the input image size. \textcolor{black}{Rather, it} changes the network \textcolor{black}{after} every few iterations. \textcolor{black}{After}, every ten batches, the network randomly chooses a new image dimension size, and the training is resumed. This forces the network to learn to \textcolor{black}{accurately} predict across a variety of input dimensions. In Section \ref{input-image-size:sec}, we show through experiments that during recognition, the input size can be easily defined as a function of the testing input image. Because Yolo looks at the whole input, it implicitly encodes contextual information about objects and their neighborhood.

\begin{figure}[h!]
	\centering
	\epsfig {file=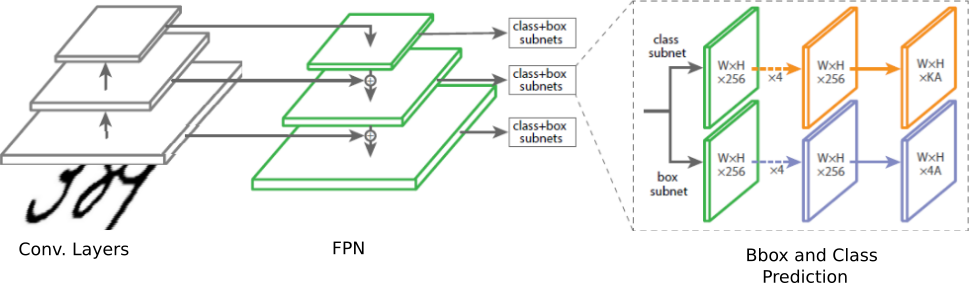, width=.95\textwidth}
	\caption{RetinaNet Framework: A Feature Pyramid Network (FPN) on top of convolutional layers produces rich and multi-scale features from one single input. Moreover, the proposed loss function (\textit{focal}) improved the class imbalance issue among background and foreground samples.}
	\label{retinaframework:fig}
\end{figure}

The RetinaNet \textcolor{black}{architecture} (\cite{RetinaNet2017}) is depicted in Figure \ref{retinaframework:fig}. A Feature Pyramid Network (FPN) on the top of convolutional layers produces rich and multi-scale features based on a single input resolution. \textcolor{black}{Compared with} Yolo, both frameworks have a similar workflow \textcolor{black}{despite these slight changes}:  the convolutional layers produce features to bounding box regressors and class predictors, \textcolor{black}{which}, with the aid of anchors boxes, determine locations and classes for objects in the input image. \textcolor{black}{\ref{YoLo_layers:sec} provides} detailed information about convolutional layers \textcolor{black}{as well as} a definition of anchors.

\textcolor{black}{What distingues RetinaNet from other approaches is its} proposed loss function, \textcolor{black}{also know as the} \emph{focal loss}. The authors evidence that a \textcolor{black}{significant} issue \textcolor{black}{encountered} in most object detection approaches is the class imbalance \textcolor{black}{that exists} among foreground and background samples. Since most image locations do not contain an object of interest, the ratio between foreground and background locations is about 1:100 or even 1:1000. \textcolor{black}{Therefore}, the background samples dominate the loss gradient, \textcolor{black}{and consequently}, the result is a biased model. The solution proposed is to define a loss function that penalizes ``easy'' classified samples. 

Let the cross-entropy loss (CE) for classification be:

\eqnnm{ce}{\CE(p,y) = \begin{cases} -\log(p) &\text{if $y = 1$}\\
		-\log (1 - p) &\text{otherwise.}\end{cases}}

\noindent where $y \in \{\pm1\}$ denotes the ground-truth class and $p \in [0,1]$ is the estimated probability for the class with label $y=1$. For the sake of simplicity, let $\pt$ be:
\eqnnm{pt}{\pt=\begin{cases} p &\text{if $y = 1$}\\ 1 - p &\text{otherwise,}\end{cases}}

Finally, $\CE(p,y) = \CE(\pt) = - \log (\pt)$.

Once a weighting factor ($- \at \log (\pt)$) should balance the priority of background and foreground, it does not give attention to easy or hard samples. \textcolor{black}{Therefore}, the author proposes to add a modulating factor {\small $(1 - \pt)^\gamma$} to the cross-entropy loss, with tunable focusing parameter $\gamma \ge 0$:

\vspace{-3mm}
\eqnnm{flalpha}{\FL(\pt) = - \at (1 - \pt)^\gamma \log (\pt).}
\vspace{-6mm}

When an example is misclassified and $\pt$ is small, the modulating factor is close to $1$, and the loss is unaffected. As $\pt \rightarrow 1$, the factor goes to 0 and the loss for well-classified examples is down-weighted. The focusing parameter $\gamma$ smoothly adjusts the rate at which easy examples are down-weighted. When $\gamma = 0$, FL is equivalent to CE, and as $\gamma$ is increased, the effect of the modulating factor is likewise increased.

\subsubsection{Network Architectures}\label{YoLo_layers:sec}

The network architectures used by both Yolo and RetinaNet are presented in Table \ref{convlayers:tab}. Yolo was first introduced with an architecture called Darknet (\cite{YOLO2017}) to perform the classification of 1000 object categories. It is composed of 19 convolutional layers and 5 max-pooling layers. To perform detection, they suppressed the last convolutional layer and added three $3 \times 3$ convolutional layer with 1024 filters. 

The concept of residual networks (ResNet) was introduced by \cite{ResNet2016} to deal with the vanish gradient issue in deep networks. \textcolor{black}{It provided a breakthrough as it allowed to skipping} connections between convolution blocks. Using this concept, the authors proposed several networks between 34 and 152 layers, \textcolor{black}{and which} achieved outstanding performance on the benchmark datasets. \textit{The ResNet-50 provides} a good tradeoff between speed and accuracy and it is the backbone for the RetinaNet framework. \textcolor{black}{Its architecture is detailed in  Table \ref{convlayers:tab}}. Moreover, an FPN with levels \textcolor{black}{ranging} from $P_3$ to $P_7$, produces rich and multi-scale features from a single input resolution.

\newcommand{\blockb}[3]{\multirow{3}{*}{\(\left[\begin{array}{c}\text{1$\times$1, #2}\\[-.1em] \text{3$\times$3, #2}\\[-.1em] \text{1$\times$1, #1}\end{array}\right]\)$\times$#3}
}

	\begin{table}[h!]
	\caption{Architectures of Darknet (left) and ResNet-50 (right). In the ResNet-50, a downsampling with a stride of 2 is performed after each convolutional block.}
	\begin{center}

	\begin{tabular}{cc}
		
		\begin{minipage}{.5\linewidth}
		\scalebox{0.93}{
			\begin{tabular}{cccc}
				\multicolumn{4}{c}{\textbf{Darknet (Yolo)}} \\
				\hline  
				\textbf{Layer} & \textbf{Type} & \textbf{Filters} & \textbf{Size/Stride} \\	\hline
				\#1 & Conv. & 32 & 3x3 / 1 \\
				\#2 & Maxpool &  & 2x2 / 2 \\
				\#3 & Conv. & 64 & 3x3 / 1 \\
				\#4 & Maxpool &  & 2x2 / 2 \\
				\#5 & Conv. & 128 & 3x3 / 1 \\
				\#6 & Conv. & 64 & 1x1 / 1 \\
				\#7 & Conv. & 128 & 3x3 /1 \\
				\#8 & Maxpool &  & 2x2 / 2 \\
				\#9 & Conv. & 256 & 3x3 / 1 \\
				\#10 & Conv. & 128 & 1x1 / 1 \\
				\#11 & Conv. & 256 & 3x3 / 1 \\
				\#12 & Maxpool &  & 2x2 / 2 \\
				\#13 & Conv. & 512 & 3x3 / 1 \\
				\#14 & Conv. & 256 & 1x1 / 1 \\
				\#15 & Conv. & 512 & 3x3 / 1 \\
				\#16 & Conv. & 256 & 1x1 / 1 \\
				\#17 & Conv. & 512 & 3x3 / 1 \\
				\#18 & Maxpool &  & 2x2 / 2 \\
				\#19 & Conv. & 1024 & 3x3 / 1 \\
				\#20 & Conv. & 512 & 1x1 / 1 \\
				\#21 & Conv. & 1024 & 3x3 / 1 \\
				\#22 & Conv. & 512 & 1x1 / 1 \\
				\#23 & Conv. & 1024 & 3x3 / 1 \\ \hline
				\#24 & Conv. & 1000 & 1x1 \\
				\#25 & Avgpool &  & Global\\
				\#26 & Softmax &  &  \\ \hline
			\end{tabular}}
		\end{minipage}
		&	

			\begin{minipage}{.5\linewidth}
			\vspace{-38.5mm}
			\scalebox{0.93}{
				\begin{tabular}{ccc}
					\multicolumn{3}{c}{\textbf{ResNet-50 (RetinaNet)}} \\
					\hline
					\textbf{Layer} & \textbf{Type} & \textbf{Filters}  \\
					\hline
					\#1 & Conv. & 7$\times$7, 64, stride 2\\

					\#2 & Max-Pool & 3$\times$3, stride 2\\

					\multirow{3}{*}{\#3..11} & \multirow{3}{*}{Conv.} & {\blockb{256}{64}{3}} \\
					&  & \\
					&  & \\

					\multirow{3}{*}{\#12..23} & \multirow{3}{*}{Conv.} & {\blockb{512}{128}{4}} \\
					&  & \\
					&  & \\					

					\multirow{3}{*}{\#24..41} & \multirow{3}{*}{Conv.} & {\blockb{1024}{256}{6}} \\
					&  & \\
					&  & \\					

					\multirow{3}{*}{\#42..50} & \multirow{3}{*}{Conv.} & {\blockb{2048}{512}{3}} \\
					&  & \\
					&  & \\					
					\hline
					\#51 &  & Avgpool \\
					\#52 & & 1000-d FC \\
					\#53 & & Softmax \\		
					\hline

				\end{tabular}
			}
			\end{minipage} 

	\end{tabular}

	\end{center}
	\label{convlayers:tab}
	\end{table}

The default dimensions of anchor boxes were defined by authors using samples of the Imagenet Dataset, composed of 1000 classes of real-life objects. Although \textcolor{black}{the dataset} includes a wide range of classes, to make anchors feasible for digits, we performed a $k$-means clustering over 10,000 ground-truth bounding boxes from the training samples. \textcolor{black}{This resulted in} three anchors with \textcolor{black}{the following} aspect ratios: 0.5, 0.6 and 1.0.
	
\subsection{Sequence-to-Sequence Approach}\label{crnn:sec}

\begin{figure}[h!]
		\centering
		\subfigure[] {\epsfig {file=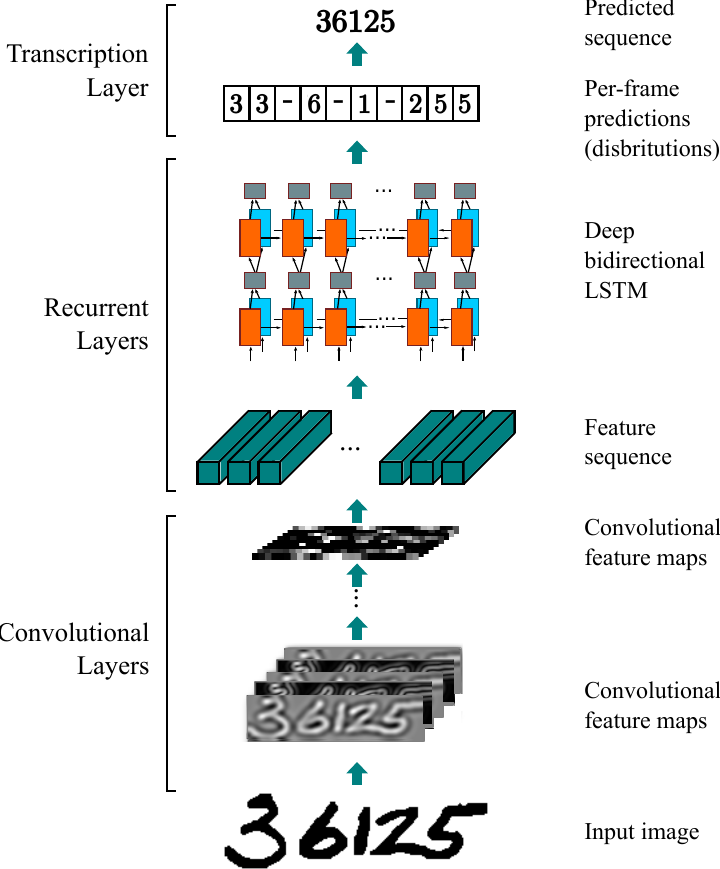, width=.59\textwidth}}
		\hspace{1cm}
		\subfigure[] {\epsfig {file=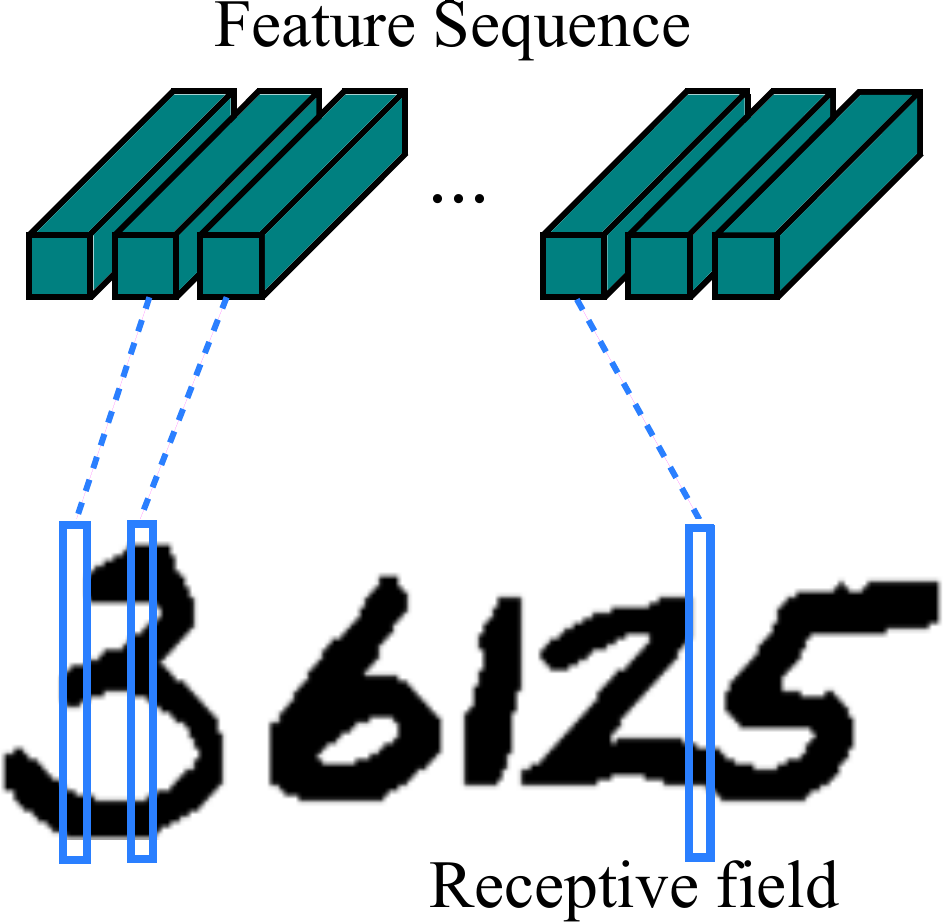, width=.25\textwidth}}
		\caption{CRNN architecture proposed by \cite{Shi2017-CRNN}: (a) the pipeline from convolutional layers to transcription layer and (b) the receptive field for each feature vector.}
		\label{crnn_pipe:fig}
	\end{figure}

\textcolor{black}{A} Convolutional Recurrent Neural Networks (CRNN) (\cite{Voigtlaender2016,Shi2017-CRNN,Dutta2018}) is a sequence-to-sequence model \textcolor{black}{that can be trained} from end-to-end. The pipeline \textcolor{black}{for a such network} in Figure \ref{crnn_pipe:fig}a. First, convolutional layers extract features from an input image, and then a sequence of feature vectors is extracted from feature maps.

Since each region of the feature map is associated with a receptive field in the input image, each vector in the sequence is a descriptor of this image field, as illustrated in Figure \ref{crnn_pipe:fig}b. Next, this sequence fed the recurrent layers, which \textcolor{black}{are} composed of a bidirectional Long-Short Term Memory (LSTM) (\cite{Schuster1997}) network, producing a per-frame prediction from left to right of the image. Finally, the transcription layer determines the correct sequence of classes to the input image by removing the repeated adjacent labels and the blanks, represented by the character `-'. This solution is well suited when the past and future context of a sequence contribute to the recognition of the whole input. With the aid of contextual information, such as a lexicon, this approach achieves high text recognition \textcolor{black}{performance}. The application of this solution to handwritten digits is a matter of discussion once we have fewer classes than words (0..9), but there is no lexicon to mitigate possible confusion.
	
\subsubsection{Network Architecture}\label{crnn_architecture:sec}
	
 \cite{Shi2017-CRNN} proposed the CRNN architecture to recognize English words. To produce feature maps with a larger width, they adopted $1 \times 2$ size \textcolor{black}{max-pooling} on layers \#7 and \#12 instead of squared ones. The input resolution is defined as $32 \times 128$ $(height \times width)$. We kept the network architecture unchanged \textcolor{black}{where} we want to evaluate handwritten digit recognition \textcolor{black}{performance}.
	
	\begin{table}[h!t]
	\caption {CRNN Architecture proposed by \cite{Shi2017-CRNN}}
	\begin{center}
		\scalebox{0.9}{
			\begin{tabular}{cccc}
				\hline  
				\textbf{Layer} & \textbf{Type} & \textbf{Filters} & \textbf{Size/Stride}\\	\hline
				\#1 & Convolutional & 64 & 3x3 / 1\\
				\#2 & Maxpool &  & 2x2 / 2 \\
				\#3 & Convolutional & 128 & 3x3 / 1 \\
				\#4 & Maxpool &  & 2x2 / 2 \\
				\#5 & Convolutional & 256 & 3x3 / 1 \\
				\#6 & Convolutional & 256 & 3x3 / 1 \\
				\#7 & Maxpool &  & 1x2 / 2  \\
				\#8 & Convolutional & 512 & 3x3 / 1 \\
				\#9 & BatchNormalization &  &  \\
				\#10 & Convolutional & 512 & 3x3 / 1 \\
				\#11 & BatchNormalization &  & \\
				\#12 & Maxpool &  & 1x2 / 2  \\			
				\#13 & Convolutional & 512 & 2x2 / 1 \\	
				\#14 & Map-to-Sequence & & \\	
				\#15 & Bidirecional-LSTM & 256 (hidden units) &  \\					
				\#16 & Bidirecional-LSTM & 256 (hidden units) &   \\	
				\#17 & Transcription &  & \\	\hline
				\label{crnn:tab}
			\end{tabular}}
		\end{center}
	\vspace{-5mm}
	\end{table}

\subsection{Training}\label{YoLo_trainning:sec}

\begin{figure}[!htbp]
	\begin{center}
		\subfigure[] {\epsfig {file=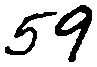, width=1.8cm, height=1.2cm}}
		\hspace{1cm}
		\subfigure[] {\epsfig {file=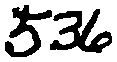, width=1.8cm, height=1.2cm}}
		\hspace{1cm}			
		\subfigure[] {\epsfig {file=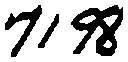, width=2cm, height=1.2cm}}						
		\hspace{1cm}			
		\subfigure[] {\epsfig {file=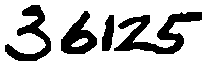, width=2.1cm, height=1.2cm}}
		\hspace{1cm}
		\subfigure[] {\epsfig {file=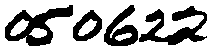, width=2.2cm, height=1.2cm}}
		\caption{Synthetic data representing numerical strings ranging from 2 to 6 digits.}
		\label{synthdataset:fig}
		\vspace{-5mm}
	\end{center}
\end{figure}

Since deep networks require a considerable amount of data to learn a representation, we created a synthetic dataset composed of numerical strings ranging from 2- to 6-digits, and containing isolated and touching components. The rationale for this strategy was to create a dataset with contextual information about the neighborhood of isolated and touching digits.  The strings are built by concatenating isolated digits of NIST SD19 (\cite{NISTSD192016}) through the algorithm described in \cite{Ribas2013}. Figure \ref{synthdataset:fig} shows some samples.

To avoid building a biased dataset, we used information on authors available on the NIST SD19, \textcolor{black}{which ensure} that digits from different authors were used exclusively for training, validation, and testing. Table \ref{tab:datadistrib} shows the purpose (training, validation, and testing), as well as the amount of data created\footnote{All the synthetic data is available upon request for research purposes at https://web.inf.ufpr.br/vri/databases-software/touching-digits/}.

\begin{table}[!htb]
	\caption {Distribution of the synthetic dataset}
	\begin{center}
		\begin{tabular}{lrll} \hline
			\multicolumn{1}{c}{Length/Classes} &
			\multicolumn{1}{c}{Samples} &
			\multicolumn{1}{c}{Authors} &
			\multicolumn{1}{c}{Purpose} \\ \hline

			2-Digit String      & 42,614 & 1000-1599 & Training \\  
			& 14,202 & 1600-1799 & Validation  \\  
			& 14,838 & 1800-1999 & Testing \\  \hline   
			3-Digit String      & 76,890 & 1000-1599 & Training \\  
			& 25,570 & 1600-1799 & Validation  \\  
			& 27,025 & 1800-1999 & Testing \\  \hline   
			4-Digit String      & 82,625 & 1000-1599 & Training \\  
			& 27,487 & 1600-1799 & Validation  \\  
			& 29,166 & 1800-1999 & Testing  \\ \hline 	
			5-Digit String      & 82,944 & 1000-1599 & Training \\  
			& 27,663 & 1600-1799 & Validation  \\  
			& 29,371 & 1800-1999 & Testing  \\ \hline 	
			
			6-Digit String      & 82,926 & 1000-1599 & Training \\  
			& 27,609 & 1600-1799 & Validation  \\  
			& 29,396 & 1800-1999 & Testing  \\ \hline

			\label{tab:datadistrib}
		\end{tabular}
	\end{center}
	\vspace{-5mm}
\end{table}

\begin{figure}[!h]
	\begin{center}
		\subfigure[] {\epsfig {file=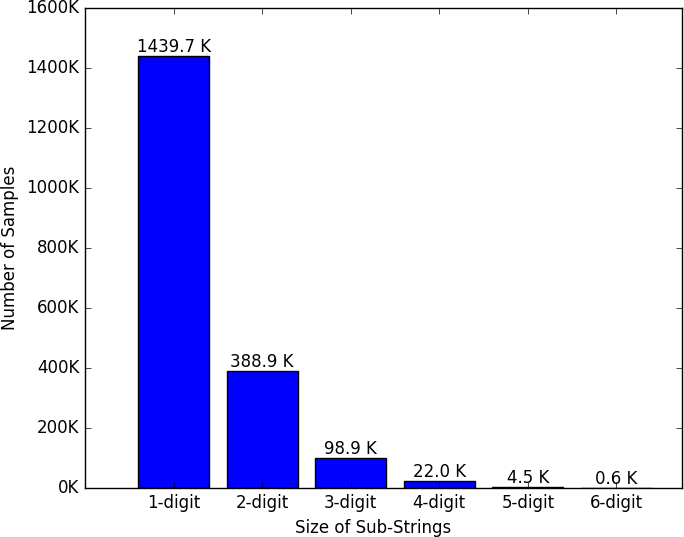, scale=0.4}}
		\hspace{1cm}
		\subfigure[] {\epsfig {file=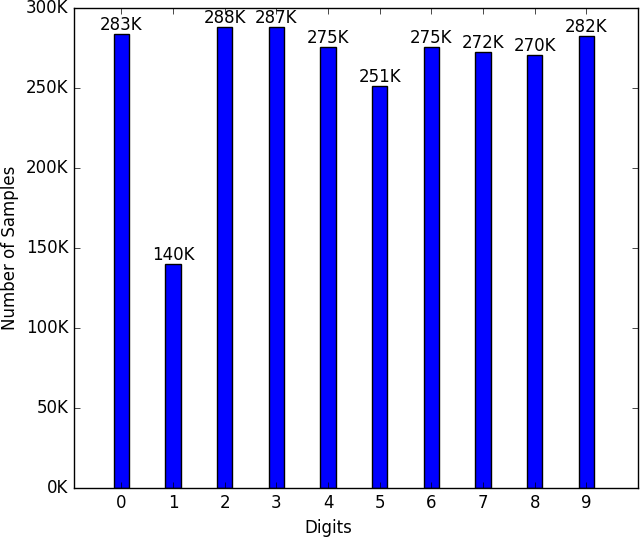, scale=0.4}}
		\caption{Distribution of the dataset (a) Distribution regarding isolated and touching digits, and (b) Distribution of the 10 classes of digits in the database.}
		\label{distrib:fig}
		\vspace{-5mm}
	\end{center}
\end{figure}

Another aspect we took into consideration when creating this dataset was the distribution of isolated and touching digits in the strings. When analyzing real datasets, one may observe something similar to an exponential distribution dominated by isolated digits. Figure \ref{distrib:fig}a shows such a distribution while \ref{distrib:fig}b depicts the distribution of the 10 classes of digits in the database. The digit ``1'' is less represented since it is the class with less occurrence in touching strings (\cite{Ribas2013}).

The models detailed in Sections \ref{ds-approach:sec}, \ref{object-detection-models:sec} and \ref{crnn:sec} were trained from scratch using the synthetic data described in Table \ref{tab:datadistrib}. Except by input size, training is performed with the Stochastic Gradient Descent (SGD) using back-propagation with mini-batches of 64 instances, a momentum factor of 0.9, and a weight decay of $5 \times 10^{-4}$. \textcolor{black}{Initially}, the learning rate is set to $10^{-3}$, to allow the weights to quickly fit the long ravines in the weight space, after which it is reduced over time (until  $5 \times 10^{-4}$) to make the weights fit the sharp curvatures. 

In the present work, regularization was implemented through early-stopping, which prevents overfitting from interrupting the training procedure once the performance of the network on a validation set deteriorates. During training, the network's performance on the training set will continue to improve, but its performance on the validation set will only improve up to a certain point, where the network starts to overfit the training data. At that point, the learning algorithm is terminated. The models were trained using an NVidia GeForce Titan X GPU\footnote{All trained classifiers are available for research purposes at https://web.inf.ufpr.br/vri/databases-software/touching-digits/}. 

\subsubsection{Time Consuming}\label{sec:time-consuming}

Table \ref{time-consuming:tab} presents the average time consumed by each approach in terms of recognition. Since training is not often used, the impact of the time consumed for this task is not considered in this evaluation. 

In light of this, we can observe that the number of objects  (digits) \textcolor{black}{that composing} a string does not contribute to a significant increase in the recognition time for all approaches. The \textcolor{black}{reason for} this is that the network forward has a similar cost \textcolor{black}{irrespective of} the number of objects in the input. It is worth mentioning that the time analysis for \cite{Aly2019} is not reported once the code is not released.

\begin{table} [htbp]
	\caption {Average recognition time of end-to-end approaches}
	\begin{center}
		\begin{tabular}{cccc}
			\hline
			Method &  \#Models (\#Classes) & \multicolumn{2}{c}{Recognition (sec)$^1$} \\ \cline{3-4} 
			& & 1-Digit & 3-Digit\\ \hline
			CRNN & 1 (10) & 0.001 & 0.001 \\
			Yolo & 1 (10) & 0.010 & 0.011 \\
			\cite{Hochuli2018}   & 4 (1114) & 0.060 & 0.062 \\
			RetinaNet & 1 (10) & 0.160 & 0.161 \\ \hline
			\multicolumn{4}{r}{\small (1) NVIDIA Titan Xp GPU}

		\end{tabular}
		\label{time-consuming:tab}
	\end{center}
\end{table}

\section{Experiments}\label{sec:Experiments}

We designed a set of experiments on five different benchmarks to \textcolor{black}{allow a better} comparison of the different approaches. Firstly, we used the challenging Touching Pairs (TP) dataset (Section \ref{tpdexp:sec}), which contains different touching pairs \textcolor{black}{styles}. Then, we focus on the Synthetic Touching Strings dataset (Section \ref{syntheticdata_exp:sec}) to evaluate the limits of each approach in a hard task, i.e., \textcolor{black}{one using} strings with up to four touching digits. The third dataset (Section \ref{nistexp:sec}) is a well-known NIST-SD19 composed of 11,585 strings ranging from 2 to 6 digits. The fourth benchmark \textcolor{black}{was built for the ICFHR 2014 HDSR challenge} (Section \ref{orand-cvl:sec}), which contains two different datasets. Finally, we present an experiment with very long strings to emphasize the power of the object-detection approach.

\subsection{TP dataset}\label{tpdexp:sec}

The TP dataset contains 79,464 samples of touching digits and it was proposed in \cite{Ribas2013} as a benchmark for segmentation algorithms. The authors were interested in evaluating when the segmentation cuts may produce a correct segmentation no matter how many cuts were produced. The solution \textcolor{black}{in these situations is} straightforward for approaches that produce only one cut: if the resulting components (after classification) match the ground-truth, the segmentation is \textcolor{black}{deemed} correct. However, for approaches that produce multiple cuts, \textcolor{black}{the segmentation is only deemed correct}, if there are at least two correct digits among hypotheses. 

\begin{figure}[!h]
	\centering
	\epsfig{file=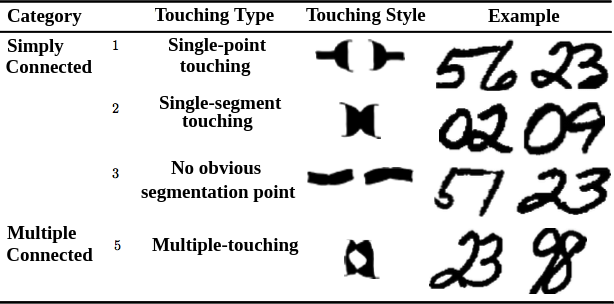, width=0.450\textwidth}
	\caption{Types of connected numeral string (extracted from \cite{Ribas2013}).}
	\label{touching:fig}
	
\end{figure}

For this experiment, we assume a correct segmentation when the model provides the correct number of digits/objects and classes. Otherwise, \textcolor{black}{there is} an error. \textcolor{black}{Two} sources of errors \textcolor{black}{are possible}: a wrong estimation of the string length or its misclassification. Table \ref{summarypairs:tab} compares the results of the end-to-end approaches with both segmentation-based and segmentation-free algorithms. \textcolor{black}{It should be mentioned} that all the works presented in Table 5 use the same testing set proposed in Ribas et al. (2013). The training sets for both the segmentation-based and the segmentation-free \textcolor{black}{algorithms used} isolated digits extracted from NIST SD19. However, \textcolor{black}{they differ in that} all segmentation-based approaches use isolated digits to train single-digit classifiers while the segmentation-free \textcolor{black}{ones} use the strings of digits described in Table 3. Table \ref{summarypairs:tab} also \textcolor{black}{illustrates} the performance according to the connection types depicted in Figure  \ref{touching:fig}.

\begin{table*} [!ht]
	\caption {Performance of the segmentation algorithms (reported in \cite{Ribas2013}, \cite{Hochuli2018}, \cite{Gattal2015}), in terms of correct segmentation, on the TP Database.}
	\begin{center}
		\scalebox{0.83}{
			\begin{tabular}{clccccccc} \hline
				\multicolumn{1}{c}{Strategy} &					
				\multicolumn{1}{c}{Method} &
				\multicolumn{1}{c}{Performance} &
				\multicolumn{4}{c}{Connection Type (\%)} &
				\multicolumn{1}{c}{Segmentation}\\ \cline{4-7}
				\multicolumn{1}{c}{} &
				\multicolumn{1}{c}{} &
				\multicolumn{1}{c}{\%} &
				\multicolumn{1}{c}{I} &
				\multicolumn{1}{c}{II} &
				\multicolumn{1}{c}{III} &
				\multicolumn{1}{c}{V} &
				\multicolumn{1}{c}{Cuts}  \\ \hline
				
				\multirow{10}{*}{\rotatebox{90}{\bf{Seg-Based}}}& \cite{Shi97}                    & 59.30 & 68.31 & 59.72 & 60.35 & 25.44 & 1    \\
				& \cite{Congedo95} 			 		 & 63.07 & 62.88 & 67.51 & 59.40 & 40.45 & 1   \\
				& \cite{Lacerda2013}   				     & 65.79 & 71.75 & 71.21 & 63.64 & 56.57 & 1    \\
				& \cite{Elnagar03}                  & 67.34 & 63.88 & 71.51 & 56.40 & 58.73 & 1    \\ 
				& \cite{Pal03}                           & 71.21 & 73.96 & 74.69 & 80.09 & 41.52 & 1    \\
				& \cite{Oliveira00-2} 	             		 & 88.03 & 90.40 & 90.78 & 89.01 & 64.88 & 1    \\
				& \cite{Fujisawa92}                         & 89.85 & 95.45 & 91.27 & 83.57 & 63.72 & 3.66 \\
				& \cite{Fenrich90}  & 92.37 & 97.54 & 93.79 & 99.45 & 65.57 & 4.07 \\
				& \cite{Gattal2015}                    & 93.24 & 96.67 & 93.75 & 99.68 & 77.58 & 24.11 \\
				& \cite{Chen00}                         & 93.80 & 97.87 & 94.23 & 97.55 & 76.76 & 45.40 \\ \hline
				\multirow{6}{*}{\rotatebox{90}{\bf{Seg-Free}}} & CRNN		 & 68.58 & 68.52 & 64.19 & 84.83 & 56.81 & 0  \\
				& RetinaNet		 & 88.48 & 89.95 & 88.51 & 97.15 & 78.32 & 0  \\
				& \cite{Aly2019}  											 & 95.05 & 95.65 & 96.20 & 97.15 & 91.21 & 0     \\
				& Yolo		 & 96.53 & 96.98 & 97.64 & 98.97 & 92.55 & 0  \\
				& \cite{Hochuli2018} 											 & 97.12 & 97.02 & 97.89 & 98.97 & 93.03 & 0     \\
				\hline
				
		\end{tabular}}
		\label{summarypairs:tab}
		
	\end{center}
\end{table*}

\vspace{-7mm}
\todo{\subsubsection{Discussion}}
Algorithms based on a single segmentation hypothesis (segmentation cuts = 1) usually fail in more complex touching cases (e.g., type V) since just one segmentation cut is often not enough to correctly split the digits. \textcolor{black}{On the other hand}, algorithms based on multiple cuts, such as \cite{Chen00, Gattal2015}, find the correct segmentation but at a high computational cost, which makes \textcolor{black}{them impractical} for real applications.

\begin{figure}[!ht]
	\begin{center}
		\subfigure[] {\epsfig {file=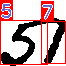, scale=0.8}}
		\hspace{2cm}
		\subfigure[] {\epsfig {file=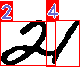, scale=0.8}}
		\hspace{2cm}
		\subfigure[] {\epsfig {file=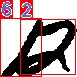, scale=0.8}}
		\hspace{2cm}
		\subfigure[] {\epsfig {file=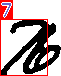, scale=0.8}}
		
		\caption{Missed detections of Yolo for TDP dataset: (a) `51' as `57', (b) `21' as `24', (c) `12' as `62' and (d) `76' as `7'.}
		\label{missed-tdp:fig}
	\end{center}
\end{figure}

Yolo compares to \cite{Hochuli2018} in terms of classification for most types of connections depicted in Figure \ref{touching:fig}, except on Type V. In this case, the task-specific classifier trained on touching pairs performs better since it can cope with highly slanted images better. This is related to the limitations of Yolo, as reported by \cite{ YOLO2016}. Yolo imposes strong spatial constraints on bounding box predictions since each grid cell only predicts two boxes and can only have one class. This spatial constraint limits the number of nearby objects that the model can predict. In our case, we observed this phenomenon in Figure \ref{missed-tdp:fig}d.

\begin{figure}[!h]
	\begin{center}
		\subfigure[] {\epsfig {file=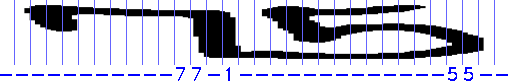, scale=0.5}}
		\hspace{1cm}
		\subfigure[] {\epsfig {file=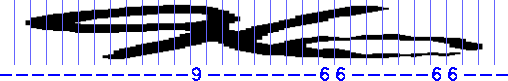, scale=0.5}}
		\hspace{1cm}
		\\
		\subfigure[] {\epsfig {file=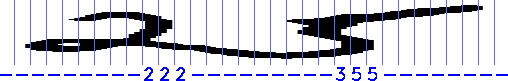, scale=0.5}}
		\hspace{1cm}
		\subfigure[] {\epsfig {file=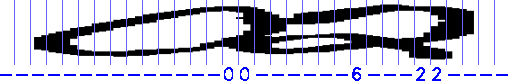, scale=0.5}}
		
		\caption{Missed predictions of CRNN for TP dataset: a) `75' as `715' (TYPE-I), b) `96' as `966' (TYPE-II), c) `25' as `235' (TYPE-III) and d) `02' as `062' (TYPE-V)  .}
		\label{crnn-tdp:fig}
	\end{center}
\end{figure}

CRNN and RetinaNet, on the other hand, performed quite poorly with performances even worse than \textcolor{black}{those of} several segmentation-based algorithms. One of the bottlenecks of the CRNN is that the local perspective of the problem given by each receptive field, or by a sub-sequence, may represent a digit fragment. In this case, a fragment of a digit taken out of context can be easily misclassified with high probability when \textcolor{black}{its shape is somewhat  similar to that of a digit}.  This issue is quite similar to the over-segmentation strategy implemented by segmentation-based approaches. Considering that there is no lexicon or post-processing method, the transcription layer may collapse by missed predictions. The worst performance is \textcolor{black}{seen} in complex cases, i.e., type V, where the neighborhood of digits is severely affected because it has more overlapping than other types. In analyzing the errors, we observe that most of \textcolor{black}{these complex} cases could be solved \textcolor{black}{using} contextual information, which, unfortunately, is not available in most applications of HDSR. \textcolor{black}{These} cases \textcolor{black}{are} depicted in Figure \ref{crnn-tdp:fig}.

\begin{figure}[!ht]
	\begin{center}
		\subfigure[] {\epsfig {file=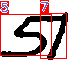, scale=0.8}}
		\hspace{10mm}
		\subfigure[] {\epsfig {file=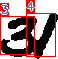, scale=0.8}}
		\hspace{10mm}
		\subfigure[] {\epsfig {file=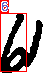, scale=0.8}}
		\hspace{10mm}
		\subfigure[] {\epsfig {file=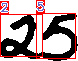, scale=0.8}}
		\hspace{10mm}
		\subfigure[] {\epsfig {file=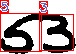, scale=0.8}}
		
		\caption{Detections of RetinaNet for TP dataset: (a) `51' as `57', (b) `31' as `34', (c) `61' as `6', representing missed prediction, and (d) `25' as `25' and e) `53' as `53' representing correct predictions.}
		\label{missed-tdp-retina:fig}
	\end{center}
\end{figure}

RetinaNet also fails \textcolor{black}{to efficiently} encode the neighborhood of digits, \textcolor{black}{which explains} the model collapse on hard overlapped digits (Type V). \textcolor{black}{It should however, be noted that it} performs well in easy cases, such as Type III. Moreover, pairs featuring the digit ``1'' produce more missed detections \textcolor{black}{if their} aspect ratio \textcolor{black}{are significantly} different from \textcolor{black}{those of the} other classes. Figure \ref{missed-tdp-retina:fig} illustrates some of these problems.

\subsection{Touching Strings Dataset}\label{syntheticdata_exp:sec}

\begin{figure}[!h]
	\begin{center}
		\subfigure[] {\epsfig {file=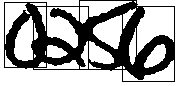, scale=0.6}}
		\hspace{0.7cm}	
		\subfigure[] {\epsfig {file=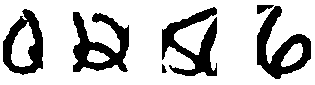, scale=0.6}}
		
		\caption{(a) Ground truth for a 4-digit string (0256) and (b) Shape of digits impacted by its neighbors.}
		\label{digit-neighboord:fig}
		\vspace{-5mm}
	\end{center}
\end{figure}

This \textcolor{black}{goal of this} experiment is to \textcolor{black}{illustrate} the limits of the evaluated approaches \textcolor{black}{when dealing with} a challenging task, i.e., \textcolor{black}{tasks involving} strings with up to four touching digits (e.g., Figure \ref{digit-neighboord:fig}a). As pointed out earlier, this is not very often observed in real databases, but it is useful for assessing the limits of the proposed strategies discussed in this work. An important point here is that, as we can observe in Figure \ref{digit-neighboord:fig}b), the shape of the digits may be severely affected the neighbors, which is quite different from those observed in the isolated digit datasets especially those in the middle of the string. This is why learning from strings rather than from isolated digits is important, particularly for approaches that use contextual information into the learning process.

In this experiment, 570,000 images of isolated digits, 2-, 3-, and 4-touching digits described in \cite{Hochuli2018} were used. The accuracy of all the strategies \textcolor{black}{employed} and the average recognition time are reported in Table \ref{summarysynthetic:tab}. As stated in Section \ref{sec:time-consuming}, a more in-depth analysis of \cite{Aly2019}'s approach is not reported as the code \textcolor{black}{was} not released.

\begin{table} [htbp]
	\caption {Accuracy of the segmentation-free approaches on the synthetic data. (The best performances are highlighted in bold)}
	\begin{center}
		\begin{tabular}{lcccc} \hline
			\multicolumn{1}{c}{Method} &
			\multicolumn{1}{c}{Isolated digit} &
			\multicolumn{1}{c}{2-digit} &
			\multicolumn{1}{c}{3-digit} &			
			\multicolumn{1}{c}{4-digit} \\ \hline

     		 \cite{Hochuli2018}  		 & \textbf{99.56} & \textbf{99.00} & 94.88 & - \\
			 {CRNN}  &  {21.97}  &   {65.33} &  {84.29} &  {90.61}  \\
			 {RetinaNet}  &  {86.63}  &   {87.32} &  {81.58} &  {77.52}  \\
			 {Yolo}  &  {99.42}  &   {98.68} &  \textbf{96.89} &  \textbf{95.50}  \\ \hline
			 
			
		\end{tabular}
		\label{summarysynthetic:tab}
	\end{center}
\end{table}

\vspace{-12mm}
\todo{\subsubsection{Discussion}}
As can be observed, the best overall results were achieved by Yolo followed by the approach proposed in \cite{Hochuli2018}. Yolo's main advantage is that it has no constraints regarding the number of touching digits in the string. 

\begin{figure}[!ht]
		\begin{center}
			\subfigure[] {\epsfig {file=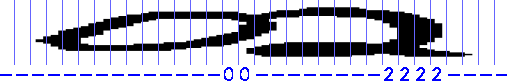, scale=0.45}}
			\hspace{1cm}
			\subfigure[] {\epsfig {file=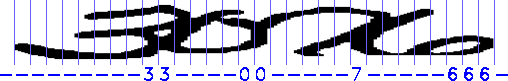, scale=0.45}}
			\hspace{1cm}
			\\
			\subfigure[] {\epsfig {file=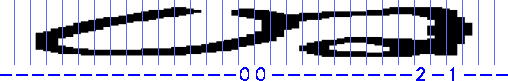, scale=0.45}}
			\hspace{1cm}
			\subfigure[] {\epsfig {file=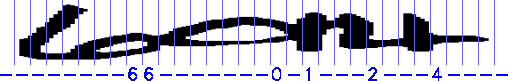, scale=0.45}}
			\\
			\subfigure[] {\epsfig {file=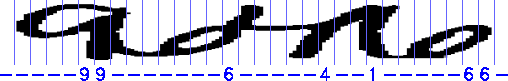, scale=0.45}}
			
			\caption{Predictions of sequence-to-sequence approach: a) `02' as `02' and b) `3076` as `3076' representing correct predictions, c) `02' as `021', d) `6014' as `60124' and e) `9646' as `96416' representing missed predictions .}
			\label{crnn-synthetic:fig}

		\end{center}
	\end{figure}

Regarding the CRNN, the design of its architecture imposes few constraints over its performance on digit strings. Since its input size is fixed, a shorter string has its aspect ratio stretched, which has more probability of suffering from over-segmentation. Figure \ref{crnn-synthetic:fig}d \textcolor{black}{illustrates} a missed prediction of digit `2' as a fragment of digit `4'. In such a case, \textcolor{black}{taking} the representation \textcolor{black}{contained} in the receptive fields, out of context, could reasonably leads to a digit `2' \textcolor{black}{being composed}. An \textcolor{black}{extrapolation} is possible \textcolor{black}{to the} missed predictions of digit `1' in Figure \ref{crnn-synthetic:fig}c and Figure \ref{crnn-synthetic:fig}e. Furthermore, in Figure \ref{crnn-synthetic:fig}, we can observe the impact of the aspect ratio and aforementioned over-segmentation.

	\begin{figure}[!h]
		\begin{center}
			\subfigure[] {\epsfig {file=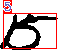, scale=1}}
			\hspace{9mm}
			\subfigure[] {\epsfig {file=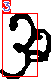, scale=1}}
			\hspace{9mm}
			\subfigure[] {\epsfig {file=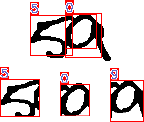, scale=1}}
		    \hspace{9mm}			
			\subfigure[] {\epsfig {file=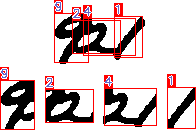, scale=1}}	
			
			\caption{Missed predictions of RetinaNet: a) `15' as `5',  b) `32` as `3',  c) `59' as `509' and d) `921' as `9241'.}
			\label{retinanet-synthetic:fig}
			
		\end{center}
	\end{figure}

\begin{figure}[!ht]
	\begin{center}
		\subfigure[] {\epsfig {file=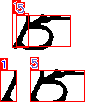, scale=0.9}}
		\hspace{5mm}
		\subfigure[] {\epsfig {file=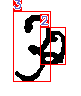, scale=0.75}}
		\hspace{5mm}
		\subfigure[] {\epsfig {file=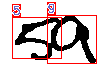, scale=0.75}}
		\hspace{5mm}
		\subfigure[] {\epsfig {file=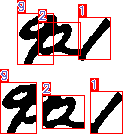, scale=0.9}}
		
		\caption{Correct predictions of Yolo approach: a) `15' as `15' , b) `32` as `32', c) `59' as `59' and d) `921' as `921'.}
		\label{Yolo-synthetic:fig}
		
	\end{center}
\end{figure}

RetinaNet suffers when encoding the neighborhood of  digit. In Figures \ref{retinanet-synthetic:fig}a and \ref{retinanet-synthetic:fig}b the aspect ratio of digits `1' and `2' are quite different from \textcolor{black}{that of the} neighborhood, \textcolor{black}{which then} results in a misclassification. In  Figure \ref{retinanet-synthetic:fig}d,  a segment touching misleads the network in the detection of a digit `4'.  Moreover, the multi-scale strategy can magnify a fragment that can be confused with a digit. In such a case, the number `9' was recognized as `0', which is quite similar to over-segmentation. Figure \ref{retinanet-synthetic:fig}c \textcolor{black}{illustrates the problem}. \textcolor{black}{The Yolo approach (Figure \ref{Yolo-synthetic:fig}) sucessfully overcomes these issues}.

\subsection{NIST SD19 Strings}\label{nistexp:sec}
	
Experiments using real-world strings are based on 11,585 numeral strings extracted from the hsf\_7 series and distributed into five classes: 2\_digit (2,370), 3\_digit (2,385) 4\_digit (2,345), 5\_digit (2,316), and 6\_digit (2,169) strings, respectively. The strings were cropped from original samples leaving a border of 5 pixels. These data exhibit different problems, such as touching and fragmentation, and were also used as test sets in \cite{Hochuli2018, Oliveira02b, Alceu2001, Liu2004, Oliveira2004, Sadri2007, Gattal2017}. It is important to mention that hsf\_7 was never used for training. 

\vspace{-7mm}
\todo{\subsubsection{Discussion}}
To better compare the approaches, we divided the errors into two classes: misdetection and misclassification. Table \ref{summary-nist-detailed:tab} summarizes the results for this experiment for the approaches.

\begin{table} [htbp]
	\caption {Recognition rates for 2- to 6-digit strings of NIST SD19 dataset}
	\begin{center}
		\begin{tabular}{lccc}
			\hline
			Method &  Recognition & \multicolumn{2}{c}{Error (\%)} \\ \cline{3-4} 
			& Rate (\%) & Classification & Detection \\ \hline
			Yolo & 97.1 & 2.4 & 0.5 \\
			\cite{Aly2019}   & 96.1 & N/A & N/A \\
			\cite{Hochuli2018}   & 95.2 & 3.9 & 0.9 \\
			CRNN & 80.3 & 11.8 & 7.9 \\
			RetinaNet & 75.3 & 1.5 & 23.2 \\\hline
		\end{tabular}
		\label{summary-nist-detailed:tab}
	\end{center}
\end{table}

\begin{figure}[!h]
		\begin{center}
			\subfigure[] {\epsfig {file=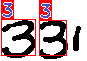, scale=0.5}}
			\hspace{0.5cm}	
			\subfigure[] {\epsfig {file=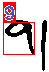, scale=0.5}}
			\hspace{0.5cm}	
			\subfigure[] {\epsfig {file=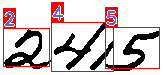, scale=0.5}}
			\hspace{0.5cm}	
			\subfigure[] {\epsfig {file=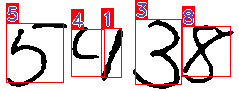, scale=0.5}}
			\\
			\hspace{0.5cm}	
			\subfigure[] {\epsfig {file=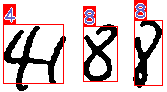, scale=0.5}}	
			\hspace{0.5cm}	
			\subfigure[] {\epsfig {file=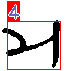, scale=0.5}}	
			\hspace{0.5cm}	
			\subfigure[] {\epsfig {file=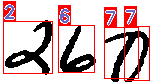, scale=0.5}}

			\caption{Detection problems: (a) 331 recognized as 33, (b) 91 recognized as 9, (c) 2415 recognized as 245, (d) 5438 recognized as 54138, (e) 4188 recognized as 488, (f) 21 recognized as 4, and (g) 260 recognized as 2670.}
			\label{misd:fig}
			\vspace{-5mm}
		\end{center}
	\end{figure}

The Yolo error analysis shows that most detection problems are related to the digit ``1''. The problem occurs when i) the height of the image is too small (Figure \ref{misd:fig}a), ii) is too high (Figure \ref{misd:fig}b) or iii) the slant of the image is big (Figure \ref{misd:fig}c). In these cases, the digit ``1'' is not detected. Another source of error is the digit ``4'' (very often related to the digit ``1''). In these cases, the model sometimes detects two objects (``4'' and ``1'' ) in the digit ``4'' (Figure \ref{misd:fig}d) and sometimes just the digit ``4'' is detected, missing the digit ``1' (Figure \ref{misd:fig}e). Finally, we observed a few samples behaving similarly to under-segmentation (Figure \ref{misd:fig}f) and over-segmentation  (Figure \ref{misd:fig}g).

\begin{figure}[!h]
		\begin{center}
			\subfigure[] {\epsfig {file=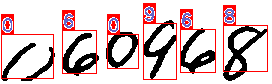, scale=0.5}}
			\hspace{0.5cm}	
			\subfigure[] {\epsfig {file=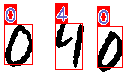, scale=0.5}}
			\hspace{0.5cm}	
			\subfigure[] {\epsfig {file=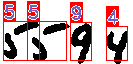, scale=0.5}}
			\hspace{0.5cm}	
			\subfigure[] {\epsfig {file=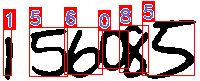, scale=0.5}}

			\caption{Correct detection: (a) 060968, (b) 040, (c) 5594, and (d) 156085}
			\label{detectionok:fig}
			\vspace{-5mm}
		\end{center}
	\end{figure}

It is worth mentioning that the average misdetection rate \textcolor{black}{was} below 1\%, and most of the cases featuring broken digits (Figures \ref{detectionok:fig}a, \ref{detectionok:fig}b, and \ref{detectionok:fig}c) and densely connected strings (Figure \ref{detectionok:fig}d), where other approaches show their \textcolor{black}{limitations}, were successfully recognized by the Yolo.

\begin{figure}[!h]
		\begin{center}
			\subfigure[] {\epsfig {file=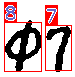, scale=0.5}}
			\hspace{0.5cm}	
			\subfigure[] {\epsfig {file=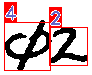, scale=0.5}}
			\hspace{0.5cm}	
			\subfigure[] {\epsfig {file=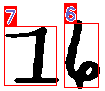, scale=0.5}}
			\hspace{0.5cm}	
			\subfigure[] {\epsfig {file=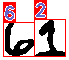, scale=0.5}}
			\hspace{0.5cm}	
			\subfigure[] {\epsfig {file=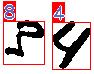, scale=0.5}}			
			\hspace{0.5cm}	
			\subfigure[] {\epsfig {file=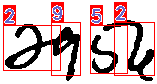, scale=0.5}}	
			
			\caption{Misclassification (a) 07 recognized as 87, (b) 02 recognized as 42, (c) 16 recognized as 76, (d) 61 recognized as 62, (e) 34 recognized as 84, and (f) 2956 recognized as 2952.}
			\label{erro:fig}
			\vspace{-5mm}
		\end{center}
	\end{figure}

Table \ref{summary-nist-detailed:tab} shows an average error rate of 2.4\%, in which most \textcolor{black}{misclassifications is} related to handwriting \textcolor{black}{variability}. Figure \ref{erro:fig} shows some common mistakes involving classes ``0'' and ``1''. In these cases, the handwriting styles are poorly represented in the training set.

In the method based on dynamic selection (\cite{Hochuli2018}), misclassification is the primary source of error, \textcolor{black}{with} 1.0\% due to length classifier and 2.9\% to digit classifiers. Since \textcolor{black}{most of the} connected components in the NIST SD19 \textcolor{black}{strings} are composed of isolated digits, the 1-digit classifier is responsible for most of the connected components classification.

\begin{figure}[h]
	\centering
	\epsfig {file=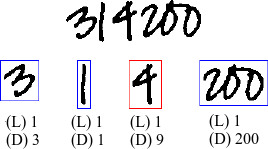, scale=0.8}
	\caption{Missed prediction of Hochuli et al.\cite{Hochuli2018}: 314200 as 319200. The classifier (L) correctly predicted the length of components, however, the 1-digit classifier (D) confused the number `4' as `9'.}
	\label{Hochuli2018:fig}
\end{figure}
	
\begin{figure}[!h]
		\begin{center}
			\subfigure[] {\epsfig {file=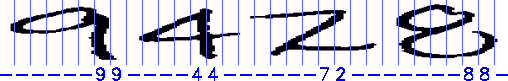, scale=0.5}}
			\hspace{0.5cm}
			\subfigure[] {\epsfig {file=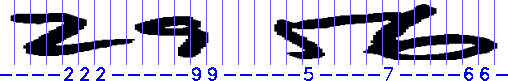, scale=0.5}}
			\hspace{0.5cm}
			\\
			\subfigure[] {\epsfig {file=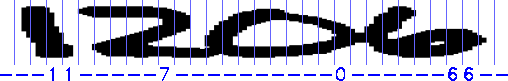, scale=0.5}}
			\hspace{0.5cm}
			\subfigure[] {\epsfig {file=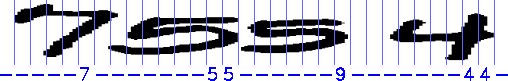, scale=0.5}}
			
			\caption{Missed predictions of CRNN for NIST dataset: a) `9428' as `94728' and b) `2956' as `29576' representing over-segmentation errors (length), and c) `1206' as '1706` and d) `7554' as `7594' representing misclassification.}
			\label{crnn-nist:fig}
		\end{center}
	\end{figure}

The detection errors of the CRNN reported in Table \ref{summary-nist-detailed:tab} occur both in isolated digits (Figure \ref{crnn-nist:fig}a) and in the touching \textcolor{black}{digits} (Figure \ref{crnn-nist:fig}b). As mentioned in Section 4.3, \textcolor{black}{the aspect ratios} of shorter and longer strings are deformed by a fixed input size, which explains the highest error rate for 2- and 6-digit strings. Performance was severely impacted by misclassification into all string sizes. Since the handwriting \textcolor{black}{was highly variable}, CRNN did not generalize the representation. This issue is depicted in Figure \ref{crnn-nist:fig}c and Figure \ref{crnn-nist:fig}d, \textcolor{black}{where the} digits `2' and  `5' were missed.

\begin{figure}[!h]
		\begin{center}
			\subfigure[] {\epsfig {file=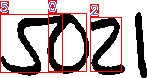, scale=0.8}}
			\hspace{0.5cm}	
			\subfigure[] {\epsfig {file=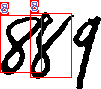, scale=0.8}}
			\hspace{0.5cm}	
			\subfigure[] {\epsfig {file=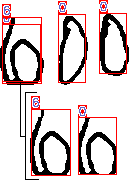, scale=0.8}}
			\hspace{0.5cm}	
			\subfigure[] {\epsfig {file=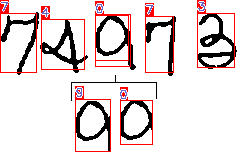, scale=0.8}}

			\caption{Missed detections of RetinaNet for NIST dataset: (a) `5021' as `502' , (b) `889' as `88', (c) `600' as `6000' and (d) `74973' as `749073' }
			\label{retina-erros:fig}
			\vspace{-5mm}
		\end{center}
	\end{figure}

Finally, Table \ref{summary-nist-detailed:tab} shows that the bottleneck of RetinaNet is detection, \textcolor{black}{as it either misdetects or overdetects digits}. The former is related to the shape of digit, while the latter is caused by a multi-scale technique which allows a fragment of a digit to be magnified to a scale that represents a digit, with high accuracy. This issue is similar to over-segmentation. The aforementioned issues are depicted in Figure \ref{retina-erros:fig}.

Table \ref{comparisonstrings:tab} compares the recognition rates of several systems reported in the literature on NIST-SD19. For completeness, we replicate the results compiled by \cite{Hochuli2018}. The works by \cite{Alceu2001}, \cite{Oliveira02b}, and \cite{Oliveira2004} use different segmentation (implicit and explicit) and classification strategies, such as Hidden Markov Models, Multi-layer Perceptrons and Support Vector Machines. Except for \cite{Liu2004} and \cite{Ciresan2008}, all the works use the same strings for testing. Regarding the training data, all of them used isolated digits from NIST SD19. However, the number of digits and how they are used may vary according to the strategy used in each system. In the case of \textcolor{black}{the} Yolo, RetinaNet, CRNN, and  \cite{Hochuli2018} approaches, the classifiers were trained  with the synthetical strings reported in Table \ref{tab:datadistrib}, which were built by combining the same isolated digits.

The work presented by \cite{Sadri2007} is reported in two columns. The authors proposed a system based on over-segmentation, in which they used a genetic algorithm to optimize their segmentation algorithm. As pointed out in \cite{Hochuli2018}, the second set of experiments (marked with an * in Table \ref{comparisonstrings:tab}) is somehow biased since the heuristics were defined using a subset of the testing set. \cite{ Gattal2017} also reported good performance, \textcolor{black}{but evaluating} their results is c\textcolor{black}{complicated by the fact} that several thresholds used for segmentation appear to be adjusted on the testing set. 

Finally, a straightforward comparison is possible with the segmentation-free methods proposed in \cite{Hochuli2018} and recently improved by \cite{Aly2019}, which implemented a different fusion strategy, \textcolor{black}{even while}, keeping the pre-processing steps and specific-task classifiers.  As discussed in Section \ref{hdsr_app:sec}, the end-to-end approaches cuts off all the heuristics used for pre-processing, \textcolor{black}{the need to train} several deep learning models, and the parameter used in the fusion strategy. \todo{Additionaly, Yolo improves the average recognition rate.}

\begin{table} [h!]
	\caption {Comparison of the recognition rates on NIST SD19}
	\begin{center}
		\begin{tabular}{cc|ccccccccccc|ccc} 
			\multicolumn{1}{c}{\rotatebox{90}{Length}} &
			\multicolumn{1}{c}{\rotatebox{90}{Samples}} &
			\multicolumn{1}{c}{\rotatebox{90}{RetinaNet}} &
			\multicolumn{1}{c}{\rotatebox{90}{CRNN}} &
			\multicolumn{1}{c}{\rotatebox{90}{\cite{Alceu2001}}} &
			\multicolumn{1}{c}{\rotatebox{90}{\cite{Oliveira02b}}} &
			\multicolumn{1}{c}{\rotatebox{90}{\cite{Oliveira2004}}} &
			\multicolumn{1}{c}{\rotatebox{90}{\cite{Sadri2007}}} &
			\multicolumn{1}{c}{\rotatebox{90}{*\cite{Sadri2007}}} &
			\multicolumn{1}{c}{\rotatebox{90}{\cite{Gattal2017}}} &
			\multicolumn{1}{c}{\rotatebox{90}{\cite{Hochuli2018}}} &
			\multicolumn{1}{c}{\rotatebox{90}{\cite{Aly2019}}} &
			\multicolumn{1}{c}{\rotatebox{90}{Yolo}} &
			\multicolumn{1}{c}{\rotatebox{90}{Samples}} &
			\multicolumn{1}{c}{\rotatebox{90}{\cite{Liu2004}}} &       
			\multicolumn{1}{c}{\rotatebox{90}{\cite{Ciresan2008}}} 
			\\ \hline

			2 & 2370 & 85.3 & 70.3 & 94.8 & 96.8 & 97.6 & 95.5 & 98.9 & 99.0 & 97.6 & 98.8 & 98.6 &     &      & \\
			3 & 2385 & 81.5 & 84.4 & 91.6 & 95.3 & 96.2 & 91.4 & 97.2 & 97.3 & 96.2 & 96.4 & 97.6 & 1476 & 96.8 & 93.4 \\
			4 & 2345 & 75.7 &86.8 & 91.3 & 93.3 & 94.2 & 91.0 & 96.1 & 96.5 & 94.6 & 95.0 & 97.1 &     &      &          \\
			5 & 2316 & 68.5 &83.8 & 88.3 & 92.4 & 94.0 & 88.0 & 95.8 & 95.9 & 94.1 & 95.4 & 96.5 &     &      &          \\
			6 & 2169 & 65.7 &76.3 & 89.0 & 93.1 & 93.8 & 88.6 & 96.1 & 96.6 & 93.3 & 95.0 & 95.8 & 1471 & 96.7 & \\ \hline
			\multicolumn{2}{c|}{\textbf{Average}} & 75.3 & 80.3 & 91.0 & 94.2 & 95.2 & 90.9 & 96.8 & 97.1 & 95.2 & 96.1 & 97.1 &  & 96.7 & 93.4 \\\hline 
			
		\end{tabular}
		\label{comparisonstrings:tab}
	\end{center}
\end{table}

\subsection{ICFHR Datasets}\label{orand-cvl:sec}

The experiment \textcolor{black}{in this case} performed on two real-world datasets built for the ICFHR 2014 challenge on HDSR (\cite{Diem2014}).

\begin{figure}[!h]
	\begin{center}
		\subfigure[76210] {\epsfig {file=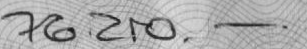, scale=0.5}}
		\hspace{0.5cm}			
		\subfigure[1455542] {\epsfig {file=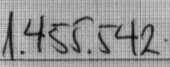, scale=0.5}}
		\hspace{0.5cm}	
		\subfigure[60000] {\epsfig {file=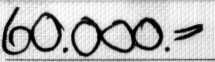, scale=0.5}}
		\hspace{0.5cm}	
		\subfigure[5841077] {\epsfig {file=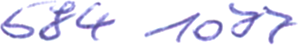, scale=0.5}}
		\hspace{0.5cm}	
		\subfigure[136075] {\epsfig {file=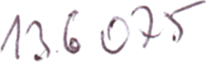, scale=0.5}}
		
		\caption{Sample data of (a) Car-A, (b) and (c) are samples of Car-B and (d) and (e) are samples of CVL dataset.}
		\label{orand-cvl:fig}
		
	\end{center}
\end{figure}

The ORAND-CAR-2014 consists of digit strings of the courtesy amount recognition (CAR) field extracted from real bank checks with a resolution of 200 dpi. Besides the traditional challenges \textcolor{black}{present} in handwriting such as noise, broken digits, and touching, this dataset presents samples with background and currency symbols such as `\#',  `\$', dots, commas, and dashes. The CVL Database was collected mostly amongst students of the Vienna University of Technology, and contains about 300 writers, female and male alike. The images are delivered with RGB information and at a resolution of 300 dpi. It includes \textcolor{black}{varying sizes and writing styles}. This database poses new challenges to the community since it is harder than previously published datasets, especially in terms of variance in writing style. Table \ref{orand-cvl-info:tab} shows the amount of data used for training and testing in both datasets. Some samples are depicted in Figure \ref{orand-cvl:fig}.

\todo{Whenever the handwriting styles of these datasets are different from those of NIST SD19, models already trained using synthetic data provide unreliable results, since the encoded information is quite different.} \textcolor{black}{We thus} trained all models using the data described in Table \ref{orand-cvl-info:tab}, since it is the protocol suggested in the ICFHR 2014 competition. We kept the training parameters \textcolor{black}{unchanged}, following that described in Section \ref{YoLo_trainning:sec}. To provide sufficient information to the object-detection approach, we annotated the digits bounding-boxes (ground-truths) of each training sample\footnote{The annotated dataset is available upon request for research purposes at https://web.inf.ufpr.br/vri/databases-software/touching-digits/}. This laborious task was necessary since most of the samples have a complex background, noise, and symbols, which are difficult to reproduce synthetically.  	

	\begin{table}[!h]
		\begin{center}
			\caption{Distribution of Orand-Car and CVL datasets}
			\begin{tabular}{c|ccc|ccc|ccc}
				\hline
				&  & Car-A &  &  & Car-B &  &  & CVL &  \\
				Length & Train & Val & Test & Train & Val & Test & Train & Val & Test \\ \hline
				2 & 17 & 5 & 36 & 0 & 0 & 0 & 0 & 0 & 0 \\
				3 & 176 & 28 & 387 & 0 & 0 & 0 & 0 & 0 & 0 \\
				4 & 633 & 71 & 1425 & 60 & 3 & 5 & 0 & 0 & 0 \\
				5 & 819 & 84 & 1475 & 1080 & 120 & 69 & 113 & 12 & 789 \\
				6 & 127 & 18 & 363 & 1432 & 167 & 1241 & 683 & 75 & 4144 \\
				7 & 27 & 2 & 87 & 127 & 10 & 1452 & 340 & 39 & 1765 \\
				8 & 1 & 1 & 11 & 1 & 0 & 157 & 0 & 0 & 0 \\
				9 & 0 & 0 & 0 & 0 & 0 & 2 & 0 & 0 & 0 \\
				10 & 0 & 0 & 0 & 0 & 0 & 0 & 0 & 0 & 0 \\ \hline
				Total & 1800 & 209 & 3784 & 2700 & 300 & 2926 & 1136 & 126 & 6698 \\ \hline
			\end{tabular}
			\label{orand-cvl-info:tab}
		\end{center}
	\end{table}

\vspace{-12mm}
\todo{\subsubsection{Discussion}}

\todo{Table \ref{orand-cvl:tab} presents the performances of end-to-end approaches on the testing set.
The performances of all methods are reported} on the same testing datasets (Table \ref{orand-cvl-info:tab}), which were proposed in the ICFHR 2014 challenge.  \cite{ Zhan2017} previously implemented the CRNN approach to these datasets; \textcolor{black}{and we therefore}, we just replicated the results. The worst results were found on the CVL dataset (26.01\%). Besides the unbalanced distribution between the training and testing sets, a short variety of string labels in the training set (only 10) do not provide an efficient representation of digit iterations into a sequence. For example, the sequence pair ``98'', which is not available in the training set, is found in two different strings of the testing set (``120398'', ``662498''). Table \ref{cvl-dist:tab} shows the poor variation of labels. Since these end-to-end models \textcolor{black}{must} learn the variability introduced by the neighborhood, this lack of samples strongly penalizes such models.

\begin{table}[!h]
	\caption {Comparison of the recognition rates on Orand and CVL datasets (ICFHR 2014 Competition)}
	\begin{center}
	\begin{tabular}{cccc}
		\hline
		Methods & CAR-A & CAR-B & CVL \\ \hline
		Tebessa I$\star$ & 37.05 & 26.62 & 59.30 \\
		Tebessa II$\star$ & 39.72 & 27.72 & 61.23 \\
		\cite{Hochuli2018} & 50.10 & 40.20 & 66.10 \\
		Singapore$\star$ & 52.30 & 59.30 & 50.40 \\
		RetinaNet & 72.51 & 69.17 & 61.06 \\
		Pernanbuco$\star$ & 78.30 & 75.43 & 58.60 \\
		Beijing$\star$ & 80.73 & 70.13 & 85.29 \\
		CRNN$\ast$ & 88.01 & 89.79 & 26.01 \\
		\cite{Saabni2016}$\star$$\dagger$ & \multicolumn{2}{c}{85.80} & - \\
		\cite{Zhan2017} & 89.75 & 91.14 & 2707 \\
		\cite{Xu2018} & 91.89 & 93.79 & 63.03 \\		
		Yolo & 96.20 & 96.80 & 84.20 \\ \hline
		\multicolumn{3}{l}{$\star$ Algorithms reported in \cite{Diem2014}} & \\
		\multicolumn{3}{l}{$\ast$ Reported by \cite{Zhan2017}} & \\
		\multicolumn{3}{l}{$\dagger$ Unified CAR-A and CAR-B datasets} & \\

	\end{tabular}

	\label{orand-cvl:tab}	
	\end{center}

\end{table}

\begin{table}[!h]
	\begin{center}
	\caption{Distribution of CVL dataset in terms of string labels variability}
	\begin{tabular}{c|cc}
		\hline
				&			& \# of Different \\
		Dataset	& Samples   & String Labels\\ \hline
		Train & 1136 & 10    \\
		Test & 6698 & 26   \\ \hline
	\end{tabular}
	\label{cvl-dist:tab}
	\end{center}
\end{table}

\textcolor{black}{Unlike in} the other benchmarks, \textcolor{black}{in which} the dynamic selection approach (\cite{Hochuli2018}) performed quite well, \textcolor{black}{it struggled in these experiments,} mostly because of its heuristic-based pre-processing module. Since ORAND-CAR provides a hard background and currency symbols, the pre-processing module collapsed when detecting connected components. It performed slightly better on the CVL dataset, which has no significant challenges in background suppression. However, the poor distribution of the training set penalized the performance of the specific-task classifiers.

\begin{figure}[!h]
	\begin{center}
		\subfigure[] {\epsfig {file=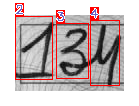, scale=0.8}}
		\hspace{0.1cm}	
		\subfigure[] {\epsfig {file=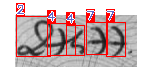, scale=0.8}}
		\hspace{0.1cm}			
		\subfigure[] {\epsfig {file=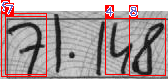, scale=0.8}}

	
		\subfigure[] {\epsfig {file=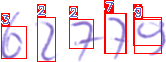, scale=1}}
		\hspace{0.7cm}			
		\subfigure[] {\epsfig {file=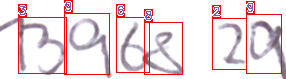, scale=1}}

		\caption{Missed predictions of RetinaNet for ORAND/CVL dataset: (a) `134' as `234' , (b) `27477' as `24477', (c) `71148' as `9748', (d) `1800000' as `800000', (e) `62779' as `32279' and (f) `1396829' as `396829'}
		\label{retina-orand:fig}
		\vspace{-5mm}
	\end{center}
\end{figure}

\begin{figure}[!h]
	\begin{center}
		\subfigure[] {\epsfig {file=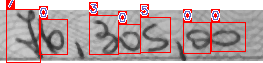, scale=0.9}}
		\hspace{0.5cm}	
		\subfigure[] {\epsfig {file=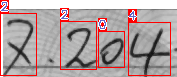, scale=0.9}}
		\hspace{0.5cm}			
		\subfigure[] {\epsfig {file=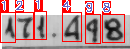, scale=0.9}}
						
		\subfigure[] {\epsfig {file=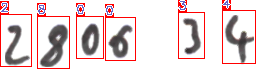, scale=0.8}}
		\hspace{0.7cm}			
		\subfigure[] {\epsfig {file=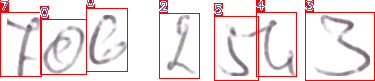, scale=0.8}}

		\caption{Missed predictions of Yolo for ORAND/CVL dataset: (a) `7630500' as `7030500' , (b) `7204' as `2204', (c) `171448' as `121498', (d) `280634' as `280034 ' and (e) `7062543' as `7002543'}
		\label{Yolo-orand:fig}
		\vspace{-5mm}
	\end{center}
\end{figure}

The \textcolor{black}{Yolo and RetinaNet} object-based models achieve a performance close to those \textcolor{black}{reported in} Section \ref{nistexp:sec}, which denote that the network could encode a hard background. A remarkable performance was achieved by the Yolo, \textcolor{black}{point to} the robustness of the model \textcolor{black}{in encoding context}, noise, and background.  The ORAND/CVL dataset also faced challenges \textcolor{black}{in the form of} overlapping digits,  handwriting variability, and different aspect ratios that severely impact the \textcolor{black}{models performances}. These issues are illustrated in Figure \ref{retina-orand:fig} and Figure \ref{Yolo-orand:fig}.

Finally, the main drawback of object-based approaches is the laborious task of data annotation when synthetic samples are not applicable.

\subsection{Very Large Strings}\label{input-image-size:sec}  

\todo{The results show} that approaching the HDSR as an object detection/recognition problem is \textcolor{black}{absolutely} feasible. \textcolor{black}{Additionaly}, it produced (with Yolo) the most consistent performance for all the benchmarks used in this study. In this final experiment, our goal is to assess the Yolo on very large strings.

As mentioned previously, the images \textcolor{black}{were} resized to $128 \times 256$ $(height \times width)$ for training. However, since Yolo changes the input size \textcolor{black}{after} every few iterations during training, this network can recognize testing images of different sizes. The question is how to properly resize the testing input image to maximize the network's performance. This is relevant since the image width may vary considerably according to the number of digits in the string. A 20-digit string is \textcolor{black}{significantly} longer than a 2-digit string, for example. Resizing both of them to $128 \times 256$ is not the right choice. 

To address this, we experimented on 5,000 strings ranging from 2 to 20 digits, which were synthetically created by concatenating isolated digits from NIST SD19. For each string length, we tested the input image width in the following range: $[128,256,384,512,640,768,896,1024,\\1152,1280]$. The image height \textcolor{black}{was} always 128. Table \ref{differentinputsize:tab} summarizes the image input size that maximizes the recognition rate for each string length. 

\begin{table} [htbp]
	\caption {Image input size that maximizes the recognition rate for each string length}
	\begin{center}
		\scalebox{0.9}{
			\begin{tabular}{cccc}
				\hline
				
				String 	& Average      &Input Image  & Recognition       \\
				Length  & String Width &Size ($II_w$)  &Rate (\%)   \\
				& ($S_w$)      &$(128 \times w)$        		  &   \\ \hline
				2 & 75  &  128 & 98.6  \\
				4 & 150 &  256 & 97.6  \\
				6 & 228 &  384 & 97.6  \\
				8 & 306 &  512 & 96.4  \\
				10 &381 &  640 & 94.8  \\
				12 &448 &  768 & 94.2  \\
				14 &524 &  896 & 91.0  \\
				16 &596 &  1024 & 90.6  \\
				18 &666 &  1152 & 88.8  \\
				20 &750 &  1280 & 89.6  \\ \hline
				
			\end{tabular}
		}
		\label{differentinputsize:tab}
		\vspace{-5mm}
	\end{center}
\end{table}

\vspace{-8mm}
\todo{\subsubsection{Discussion}}

From Table \ref{differentinputsize:tab}, we can notice that there is a quasi-linear relation between the average string width of the testing images\footnote{The number of pixels may vary depending on the image resolution. In this work, all the images were acquired in 300dpi.} and the best input size for the Yolo. In light of this, we propose a rule (Equation \ref{retina:eq}) to compute the input size width of the Yolo based on the width of the testing image. Such a rule is used for all experiments reported in this paper.

\begin{equation}
II_w = \left \{ \begin{array}{ll}
128 				& \mbox{for } S_w \leqslant 75 \\
S_w \times 1.70	& \mbox{otherwise }    \\
\end{array}
\right.
\label{retina:eq}
\end{equation}

\begin{figure}[!h]
	\begin{center}
		\subfigure[] {\epsfig {file=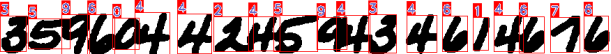, scale=0.9}}
		\\
		\subfigure[] {\epsfig {file=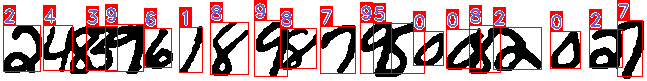, scale=.83}}
		\\
		\subfigure[] {\epsfig {file=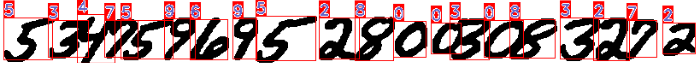, scale=0.8}}

		\caption{20-digit strings correctly recognized by the Yolo-based approach.}
		\label{longstrings:fig}
		\vspace{-5mm}
	\end{center}
\end{figure}

Figure \ref{longstrings:fig} shows some examples of 20-digit strings recognized by the system using the rule above. These corroborate the efficiency of the adopted resizing strategy and show that the approach can perform well even for very long strings composed of broken, overlapping, and different configurations of touching digits.

\subsection{Summary of the experiments}
	
\begin{figure}[!h]
		\centering
		\includegraphics[width=0.7\linewidth]{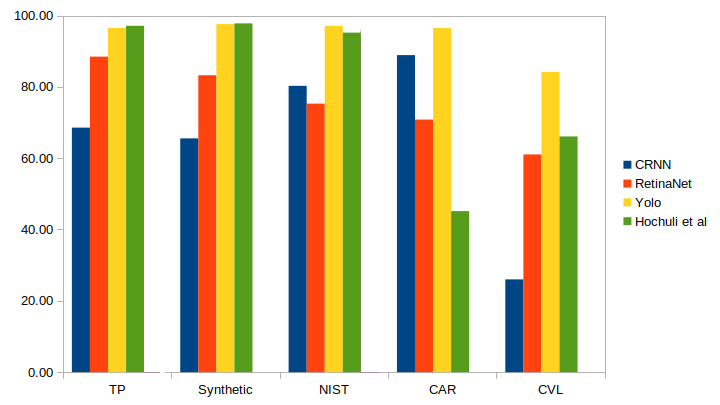}
		\caption{Average recognition of approaches per dataset}
		\label{fig:exp_summary}
	\end{figure}

Figure \ref{fig:exp_summary} summarizes the performance of the assessed methods on the different datasets used in this study.  As we can see, Yolo achieved outstanding performance in all scenarios. However, its bottleneck is the ground-truth annotation when synthetic samples are not feasible.
	
Even though RetinaNet also implements an object detection approach, it suffers from the built-in multi-scale strategy (FPN), once a magnified fragment of digit misleads the model. A similar issue occurs with CRNN in which the \textcolor{black}{various different} receptive fields fragment the input. These issues are close to the over-segmentation problem faced by segmentation-based algorithms.
	
Finally, the segmentation-free approach of \cite{Hochuli2018} perform well in scenarios where there is no hard background, \textcolor{black}{but, suffer from handling} a complex pipeline composed of heuristic process and multiple classifiers. We did not add the \cite{Aly2019} method in this comparison because we had no access to its source code.

\section{Conclusion}\label{conclusions:sec}

This paper described end-to-end solutions for HDSR in which the string of digits is assumed to be composed of objects that can be automatically detected and recognized. \textcolor{black}{To this end}, several strategies were evaluated. 

A robust experimental protocol based on numeral string datasets was defined to validate the proposed methods containing several types of noise, touching digits, fragmentation, complexes backgrounds, and long strings. The experimental results \textcolor{black}{show} that the object-detection approach is a feasible end-to-end solution that compares favorably to the state-of-the-art in HDSR \textcolor{black}{in terms of recognition rates}. Also, it considerably reduces the complexity of the string recognition task and avoiding heuristic-based methods, special pre-processing, segmentation, and classifiers devoted to specific-length strings, \textcolor{black}{meaning}, no constraints related to the string length exist. However, the \textcolor{black}{difficulty posed by need for} data annotation when synthetic samples are not applicable is the main drawback of this approach.
 
\textcolor{black}{Conversely}, the sequence-to-sequence strategy provides a short pipeline. \textcolor{black}{No significant efforts related to the annotation of ground-truth is needed, as in the case with the object-detection based approach.} However, \textcolor{black}{the strategy} depends on contextual information, such as a lexicon, to achieve good results. Thus, its design for handwritten digits needed to be reviewed. 

\section*{Acknowledgements}

This research was supported by The National Council for Scientific and Technological
Development (CNPq) grants 303252/2018-9 and 306684/2018-2, CAPES (PhD scholarship - Finance Code 001), Fondecyt Chile (project number 11150945), STIC-Amsud 19-STIC-04 and Araucária Foundation. In addition, we gratefully acknowledge the support of NVIDIA Corporation with the donation of the Titan XP GPU used for this research.

\bibliographystyle{model5-names}\biboptions{authoryear}
\bibliography{refer1}

\begin{thebibliography}{70}
\expandafter\ifx\csname natexlab\endcsname\relax\def\natexlab#1{#1}\fi
\providecommand{\url}[1]{\texttt{#1}}
\providecommand{\href}[2]{#2}
\providecommand{\path}[1]{#1}
\providecommand{\DOIprefix}{doi:}
\providecommand{\ArXivprefix}{arXiv:}
\providecommand{\URLprefix}{URL: }
\providecommand{\Pubmedprefix}{pmid:}
\providecommand{\doi}[1]{\href{http://dx.doi.org/#1}{\path{#1}}}
\providecommand{\Pubmed}[1]{\href{pmid:#1}{\path{#1}}}
\providecommand{\bibinfo}[2]{#2}
\ifx\xfnm\relax \def\xfnm[#1]{\unskip,\space#1}\fi
\bibitem[{{Aly} \& {Mohamed}(2019)}]{Aly2019}
\bibinfo{author}{{Aly}, S.}, \& \bibinfo{author}{{Mohamed}, A.}
  (\bibinfo{year}{2019}).
\newblock \bibinfo{title}{Unknown-length handwritten numeral string recognition
  using cascade of pca-svmnet classifiers}.
\newblock {\it \bibinfo{journal}{IEEE Access}\/},  {\it \bibinfo{volume}{7}\/},
  \bibinfo{pages}{52024--52034}. \DOIprefix\doi{10.1109/ACCESS.2019.2911851}.
\bibitem[{Bengio et~al.(2013)Bengio, Courville \& Vincent}]{bengio2013}
\bibinfo{author}{Bengio, Y.}, \bibinfo{author}{Courville, A.}, \&
  \bibinfo{author}{Vincent, P.} (\bibinfo{year}{2013}).
\newblock \bibinfo{title}{Representation learning: A review and new
  perspectives}.
\newblock {\it \bibinfo{journal}{IEEE Trans. on Pattern Analysis and Machine
  Intelligence}\/},  {\it \bibinfo{volume}{35}\/}, \bibinfo{pages}{1798--1828}.
\bibitem[{Britto et~al.(2014)Britto, Sabourin \& Oliveira}]{BRITTO2014}
\bibinfo{author}{Britto, A.~S.}, \bibinfo{author}{Sabourin, R.}, \&
  \bibinfo{author}{Oliveira, L.~S.} (\bibinfo{year}{2014}).
\newblock \bibinfo{title}{Dynamic selection of classifiers---a comprehensive
  review}.
\newblock {\it \bibinfo{journal}{Pattern Recognition}\/},  {\it
  \bibinfo{volume}{47}\/}, \bibinfo{pages}{3665 -- 3680}.
\bibitem[{Britto-Jr et~al.(2003)Britto-Jr, Sabourin, Bortolozzi \&
  Suen}]{Alceu2001}
\bibinfo{author}{Britto-Jr, A.}, \bibinfo{author}{Sabourin, R.},
  \bibinfo{author}{Bortolozzi, F.}, \& \bibinfo{author}{Suen, C.~Y.}
  (\bibinfo{year}{2003}).
\newblock \bibinfo{title}{The recognition of handwritten numeral strings using
  a two-stage {HMM}-based method}.
\newblock {\it \bibinfo{journal}{International Journal on Document Analysis and
  Recognition}\/},  {\it \bibinfo{volume}{5}\/}, \bibinfo{pages}{102--117}.
\bibitem[{Casey \& Lecolinet(1996)}]{Casey96}
\bibinfo{author}{Casey, R.}, \& \bibinfo{author}{Lecolinet, E.}
  (\bibinfo{year}{1996}).
\newblock \bibinfo{title}{A survey of methods and strategies in character
  segmentation}.
\newblock {\it \bibinfo{journal}{IEEE Trans. on PAMI}\/},  {\it
  \bibinfo{volume}{18}\/}, \bibinfo{pages}{690--706}.
\bibitem[{Chen \& Wang(2000)}]{Chen00}
\bibinfo{author}{Chen, Y.~K.}, \& \bibinfo{author}{Wang, J.~F.}
  (\bibinfo{year}{2000}).
\newblock \bibinfo{title}{Segmentation of single- or multiple-touching
  handwritten numeral string using background and foreground analysis}.
\newblock {\it \bibinfo{journal}{IEEE Trans. on Pattern Analysis and Machine
  Intelligence}\/},  {\it \bibinfo{volume}{22}\/}, \bibinfo{pages}{1304--1317}.
\bibitem[{Choi \& Oh(1999)}]{Choi99}
\bibinfo{author}{Choi, S.}, \& \bibinfo{author}{Oh, I.} (\bibinfo{year}{1999}).
\newblock \bibinfo{title}{A segmentation-free recognition of two touching
  numerals using neural networks}.
\newblock In {\it \bibinfo{booktitle}{Proc. of 5$^{th}$ International
  Conference on Document Analysis and Recognition}\/} (pp.
  \bibinfo{pages}{253--256}).
\newblock \bibinfo{address}{Bangalore, India}.
\bibitem[{Ciresan(2008)}]{Ciresan2008}
\bibinfo{author}{Ciresan, D.} (\bibinfo{year}{2008}).
\newblock \bibinfo{title}{Avoiding segmentation in multi-digit numeral string
  recognition by combining single and two-digit classifiers trained without
  negative examples}.
\newblock In {\it \bibinfo{booktitle}{10th International Symposium on Symbolic
  and Numeric Algorithms for Scientific Computing}\/} (pp.
  \bibinfo{pages}{225--230}).
\bibitem[{Ciresan et~al.(2012)Ciresan, Meier \& Schmidhuber}]{Ciresan2012}
\bibinfo{author}{Ciresan, D.}, \bibinfo{author}{Meier, U.}, \&
  \bibinfo{author}{Schmidhuber, J.} (\bibinfo{year}{2012}).
\newblock \bibinfo{title}{Multi-column deep neural networks for image
  classification}.
\newblock In {\it \bibinfo{booktitle}{2012 IEEE Conference on Computer Vision
  and Pattern Recognition}\/} (pp. \bibinfo{pages}{3642--3649}).
\newblock \DOIprefix\doi{10.1109/CVPR.2012.6248110}.
\bibitem[{Congedo et~al.(1995)Congedo, Dimauro, Impedovo \& Pirlo}]{Congedo95}
\bibinfo{author}{Congedo, G.}, \bibinfo{author}{Dimauro, G.},
  \bibinfo{author}{Impedovo, S.}, \& \bibinfo{author}{Pirlo, G.}
  (\bibinfo{year}{1995}).
\newblock \bibinfo{title}{Segmentation of numeric strings}.
\newblock In {\it \bibinfo{booktitle}{3rd International Conference on Document
  Analysis and Recognition}\/} (pp. \bibinfo{pages}{1038--1041}).
\bibitem[{Cruz et~al.(2018)Cruz, Sabourin \& Cavalcanti}]{CRUZ2018}
\bibinfo{author}{Cruz, R.~M.}, \bibinfo{author}{Sabourin, R.}, \&
  \bibinfo{author}{Cavalcanti, G.~D.} (\bibinfo{year}{2018}).
\newblock \bibinfo{title}{Dynamic classifier selection: Recent advances and
  perspectives}.
\newblock {\it \bibinfo{journal}{Information Fusion}\/},  {\it
  \bibinfo{volume}{41}\/}, \bibinfo{pages}{195 -- 216}.
\bibitem[{Das et~al.(2016)Das, K.Saha \& Nasipuri}]{Sarkhel2016}
\bibinfo{author}{Das, R. S.~N.}, \bibinfo{author}{K.Saha, A.}, \&
  \bibinfo{author}{Nasipuri, M.} (\bibinfo{year}{2016}).
\newblock \bibinfo{title}{A multi-objective approach towards cost effective
  isolated handwritten {B}angla character and digit recognition}.
\newblock {\it \bibinfo{journal}{Pattern Recognition}\/},  {\it
  \bibinfo{volume}{58}\/}, \bibinfo{pages}{172--189}.
\bibitem[{Diem et~al.(2014)Diem, Fiel, Kleber, Sablatnig, Saavedra, Contreras,
  Barrios \& Oliveira}]{Diem2014}
\bibinfo{author}{Diem, M.}, \bibinfo{author}{Fiel, S.},
  \bibinfo{author}{Kleber, F.}, \bibinfo{author}{Sablatnig, R.},
  \bibinfo{author}{Saavedra, J.~M.}, \bibinfo{author}{Contreras, D.},
  \bibinfo{author}{Barrios, J.~M.}, \& \bibinfo{author}{Oliveira, L.~S.}
  (\bibinfo{year}{2014}).
\newblock \bibinfo{title}{Icfhr 2014 competition on handwritten digit string
  recognition in challenging datasets (hdsrc 2014)}.
\newblock In {\it \bibinfo{booktitle}{2014 14th International Conference on
  Frontiers in Handwriting Recognition}\/} (pp. \bibinfo{pages}{779--784}).
\newblock \bibinfo{organization}{IEEE}.
\bibitem[{Divvala et~al.(2009)Divvala, Hoiem, Hays, Efros \&
  Hebert}]{divvala2009empirical}
\bibinfo{author}{Divvala, S.~K.}, \bibinfo{author}{Hoiem, D.},
  \bibinfo{author}{Hays, J.~H.}, \bibinfo{author}{Efros, A.~A.}, \&
  \bibinfo{author}{Hebert, M.} (\bibinfo{year}{2009}).
\newblock \bibinfo{title}{An empirical study of context in object detection}.
\newblock In {\it \bibinfo{booktitle}{Computer Vision and Pattern Recognition,
  2009. CVPR 2009. IEEE Conference on}\/} (pp. \bibinfo{pages}{1271--1278}).
\newblock \bibinfo{organization}{IEEE}.
\bibitem[{Dutta et~al.(2018)Dutta, Krishnan, Mathew \& Jawahar}]{Dutta2018}
\bibinfo{author}{Dutta, K.}, \bibinfo{author}{Krishnan, P.},
  \bibinfo{author}{Mathew, M.}, \& \bibinfo{author}{Jawahar, C.~V.}
  (\bibinfo{year}{2018}).
\newblock \bibinfo{title}{Improving cnn-rnn hybrid networks for handwriting
  recognition}.
\newblock In {\it \bibinfo{booktitle}{2018 16th International Conference on
  Frontiers in Handwriting Recognition (ICFHR)}\/} (pp.
  \bibinfo{pages}{80--85}).
\newblock \DOIprefix\doi{10.1109/ICFHR-2018.2018.00023}.
\bibitem[{Elms et~al.(1998)Elms, Procter \& Illingworth}]{Elms98}
\bibinfo{author}{Elms, A.~J.}, \bibinfo{author}{Procter, S.}, \&
  \bibinfo{author}{Illingworth, J.} (\bibinfo{year}{1998}).
\newblock \bibinfo{title}{The advantage of using an hmm-based approach for
  faxed word recognition}.
\newblock {\it \bibinfo{journal}{International Journal of Document Analysis and
  Recognition}\/},  (pp. \bibinfo{pages}{18--36}).
\bibitem[{Elnagar \& Alhajj(2003)}]{Elnagar03}
\bibinfo{author}{Elnagar, A.}, \& \bibinfo{author}{Alhajj, R.}
  (\bibinfo{year}{2003}).
\newblock \bibinfo{title}{Segmentation of connected handwritten numeral
  strings}.
\newblock {\it \bibinfo{journal}{Pattern Recognition}\/},  {\it
  \bibinfo{volume}{36}\/}, \bibinfo{pages}{625--634}.
\bibitem[{Felzenszwalb et~al.(2008)Felzenszwalb, McAllester \&
  Ramanan}]{Felzenszwalb2008}
\bibinfo{author}{Felzenszwalb, P.}, \bibinfo{author}{McAllester, D.}, \&
  \bibinfo{author}{Ramanan, D.} (\bibinfo{year}{2008}).
\newblock \bibinfo{title}{A discriminatively trained, multiscale, deformable
  part model}.
\newblock In {\it \bibinfo{booktitle}{Computer Vision and Pattern Recognition,
  2008. CVPR 2008. IEEE Conference on}\/} (pp. \bibinfo{pages}{1--8}).
\newblock \bibinfo{organization}{IEEE}.
\bibitem[{Felzenszwalb et~al.(2010)Felzenszwalb, Girshick, McAllester \&
  Ramanan}]{Felzenszwalb2010}
\bibinfo{author}{Felzenszwalb, P.~F.}, \bibinfo{author}{Girshick, R.~B.},
  \bibinfo{author}{McAllester, D.}, \& \bibinfo{author}{Ramanan, D.}
  (\bibinfo{year}{2010}).
\newblock \bibinfo{title}{Object detection with discriminatively trained
  part-based models}.
\newblock {\it \bibinfo{journal}{IEEE transactions on pattern analysis and
  machine intelligence}\/},  {\it \bibinfo{volume}{32}\/},
  \bibinfo{pages}{1627--1645}.
\bibitem[{Fenrich \& Krishnamoorthy(1990)}]{Fenrich90}
\bibinfo{author}{Fenrich, R.}, \& \bibinfo{author}{Krishnamoorthy, S.}
  (\bibinfo{year}{1990}).
\newblock \bibinfo{title}{Segmenting diverse quality handwritten digit strings
  in near real-time}.
\newblock In {\it \bibinfo{booktitle}{5$^{th}$ USPS Advanced Technology
  Conference}\/} (pp. \bibinfo{pages}{523--537}).
\bibitem[{Fujisawa et~al.(1992)Fujisawa, Nakano \& Kurino}]{Fujisawa92}
\bibinfo{author}{Fujisawa, H.}, \bibinfo{author}{Nakano, Y.}, \&
  \bibinfo{author}{Kurino, K.} (\bibinfo{year}{1992}).
\newblock \bibinfo{title}{Segmentation methods for character recognition: from
  segmentation to document structure analysis}.
\newblock {\it \bibinfo{journal}{Proc. of IEEE}\/},  {\it
  \bibinfo{volume}{80}\/}, \bibinfo{pages}{1079--1092}.
\bibitem[{Gattal \& Chibani(2015)}]{Gattal2015}
\bibinfo{author}{Gattal, A.}, \& \bibinfo{author}{Chibani, Y.}
  (\bibinfo{year}{2015}).
\newblock \bibinfo{title}{{SVM}-based segmentation-verification of handwritten
  connected digits using the oriented sliding window}.
\newblock {\it \bibinfo{journal}{International Journal of Computational
  Intelligence and Applications}\/},  {\it \bibinfo{volume}{14}\/},
  \bibinfo{pages}{1--17}.
\bibitem[{Gattal et~al.(2017)Gattal, Chibani \& Hadjadji}]{Gattal2017}
\bibinfo{author}{Gattal, A.}, \bibinfo{author}{Chibani, Y.}, \&
  \bibinfo{author}{Hadjadji, B.} (\bibinfo{year}{2017}).
\newblock \bibinfo{title}{Segmentation and recognition system for
  unknown-length handwritten digit strings}.
\newblock {\it \bibinfo{journal}{Pattern Analysis and Applications}\/},  {\it
  \bibinfo{volume}{20}\/}, \bibinfo{pages}{307--323}.
\bibitem[{Girshick(2015)}]{FastRCNN}
\bibinfo{author}{Girshick, R.} (\bibinfo{year}{2015}).
\newblock \bibinfo{title}{Fast r-cnn}.
\newblock In {\it \bibinfo{booktitle}{Proceedings of the 2015 IEEE
  International Conference on Computer Vision (ICCV)}\/} ICCV '15 (pp.
  \bibinfo{pages}{1440--1448}).
\newblock \bibinfo{address}{Washington, DC, USA}: \bibinfo{publisher}{IEEE
  Computer Society}.
\bibitem[{Girshick et~al.(2014)Girshick, Donahue, Darrell \& Malik}]{RCNN}
\bibinfo{author}{Girshick, R.}, \bibinfo{author}{Donahue, J.},
  \bibinfo{author}{Darrell, T.}, \& \bibinfo{author}{Malik, J.}
  (\bibinfo{year}{2014}).
\newblock \bibinfo{title}{Rich feature hierarchies for accurate object
  detection and semantic segmentation}.
\newblock In {\it \bibinfo{booktitle}{Proceedings of the IEEE conference on
  computer vision and pattern recognition}\/} (pp. \bibinfo{pages}{580--587}).
\bibitem[{Grother(2016)}]{NISTSD192016}
\bibinfo{author}{Grother, P.~J.} (\bibinfo{year}{2016}).
\newblock {\it \bibinfo{title}{NIST Special Database 19 - Handprinted forms and
  characters database}\/}.
\newblock \bibinfo{organization}{NIST}.
\bibitem[{Gu et~al.(2017)Gu, Wang, Kuen, Ma, Shahroudy, Shuai, Liu, Wang, Wang,
  Cai \& Shen}]{Gu2017}
\bibinfo{author}{Gu, J.}, \bibinfo{author}{Wang, Z.}, \bibinfo{author}{Kuen,
  J.}, \bibinfo{author}{Ma, L.}, \bibinfo{author}{Shahroudy, A.},
  \bibinfo{author}{Shuai, B.}, \bibinfo{author}{Liu, T.},
  \bibinfo{author}{Wang, X.}, \bibinfo{author}{Wang, G.}, \bibinfo{author}{Cai,
  J.}, \& \bibinfo{author}{Shen, T.} (\bibinfo{year}{2017}).
\newblock \bibinfo{title}{Recent advances in convolutional neural networks}.
\newblock {\it \bibinfo{journal}{Pattern Recognition}\/}, .
\bibitem[{Hafemann et~al.(2017)Hafemann, Sabourin \& Oliveira}]{Hafemann2017}
\bibinfo{author}{Hafemann, L.~G.}, \bibinfo{author}{Sabourin, R.}, \&
  \bibinfo{author}{Oliveira, L.~S.} (\bibinfo{year}{2017}).
\newblock \bibinfo{title}{Learning features for offline handwritten signature
  verification using deep convolutional neural networks}.
\newblock {\it \bibinfo{journal}{Pattern Recognition}\/},  (pp.
  \bibinfo{pages}{163--176}).
\bibitem[{Han et~al.(2018)Han, Zhang, Cheng, Liu \& Xu}]{Han2018}
\bibinfo{author}{Han, J.}, \bibinfo{author}{Zhang, D.}, \bibinfo{author}{Cheng,
  G.}, \bibinfo{author}{Liu, N.}, \& \bibinfo{author}{Xu, D.}
  (\bibinfo{year}{2018}).
\newblock \bibinfo{title}{Advanced deep-learning techniques for salient and
  category-specific object detection: A survey}.
\newblock {\it \bibinfo{journal}{IEEE Signal Processing Magazine}\/},  {\it
  \bibinfo{volume}{35}\/}, \bibinfo{pages}{84--100}.
  \DOIprefix\doi{10.1109/MSP.2017.2749125}.
\bibitem[{He et~al.(2017)He, Gkioxari, Doll{\'a}r \& Girshick}]{MaskRCNN}
\bibinfo{author}{He, K.}, \bibinfo{author}{Gkioxari, G.},
  \bibinfo{author}{Doll{\'a}r, P.}, \& \bibinfo{author}{Girshick, R.}
  (\bibinfo{year}{2017}).
\newblock \bibinfo{title}{Mask r-cnn}.
\newblock In {\it \bibinfo{booktitle}{Computer Vision (ICCV), 2017 IEEE
  International Conference on}\/} (pp. \bibinfo{pages}{2980--2988}).
\newblock \bibinfo{organization}{IEEE}.
\bibitem[{He et~al.(2015)He, Zhang, Ren \& Sun}]{SPPnet}
\bibinfo{author}{He, K.}, \bibinfo{author}{Zhang, X.}, \bibinfo{author}{Ren,
  S.}, \& \bibinfo{author}{Sun, J.} (\bibinfo{year}{2015}).
\newblock \bibinfo{title}{Spatial pyramid pooling in deep convolutional
  networks for visual recognition}.
\newblock {\it \bibinfo{journal}{IEEE transactions on pattern analysis and
  machine intelligence}\/},  {\it \bibinfo{volume}{37}\/},
  \bibinfo{pages}{1904--1916}.
\bibitem[{He et~al.(2016)He, Zhang, Ren \& Sun}]{ResNet2016}
\bibinfo{author}{He, K.}, \bibinfo{author}{Zhang, X.}, \bibinfo{author}{Ren,
  S.}, \& \bibinfo{author}{Sun, J.} (\bibinfo{year}{2016}).
\newblock \bibinfo{title}{Deep residual learning for image recognition}.
\newblock In {\it \bibinfo{booktitle}{Proceedings of the IEEE conference on
  computer vision and pattern recognition}\/} (pp. \bibinfo{pages}{770--778}).
\bibitem[{Hochuli et~al.(2018{\natexlab{a}})Hochuli, Oliveira, Britto \&
  Sabourin}]{Hochuli2018}
\bibinfo{author}{Hochuli, A.~G.}, \bibinfo{author}{Oliveira, L.~S.},
  \bibinfo{author}{Britto, A.~S.}, \& \bibinfo{author}{Sabourin, R.}
  (\bibinfo{year}{2018}{\natexlab{a}}).
\newblock \bibinfo{title}{Handwritten digit segmentation: Is it still
  necessary?}
\newblock {\it \bibinfo{journal}{Pattern Recognition}\/},  {\it
  \bibinfo{volume}{78}\/}, \bibinfo{pages}{1 -- 11}.
\bibitem[{Hochuli et~al.(2018{\natexlab{b}})Hochuli, Oliveira, d.~Souza~Britto
  \& Sabourin}]{Hochuli2018b}
\bibinfo{author}{Hochuli, A.~G.}, \bibinfo{author}{Oliveira, L.~S.},
  \bibinfo{author}{d.~Souza~Britto, A.}, \& \bibinfo{author}{Sabourin, R.}
  (\bibinfo{year}{2018}{\natexlab{b}}).
\newblock \bibinfo{title}{Segmentation-free approaches for handwritten numeral
  string recognition}.
\newblock In {\it \bibinfo{booktitle}{2018 International Joint Conference on
  Neural Networks (IJCNN)}\/} (pp. \bibinfo{pages}{1--8}).
\bibitem[{Lacerda \& Mello(2013)}]{Lacerda2013}
\bibinfo{author}{Lacerda, E.}, \& \bibinfo{author}{Mello, C. A.~B.}
  (\bibinfo{year}{2013}).
\newblock \bibinfo{title}{Segmentation of connected handwritten digits using
  self-organizing maps}.
\newblock {\it \bibinfo{journal}{Expert Systems with Applications}\/},  {\it
  \bibinfo{volume}{40}\/}, \bibinfo{pages}{5867--5877}.
\bibitem[{Lampert et~al.(2008)Lampert, Blaschko \& Hofmann}]{Lampert2008}
\bibinfo{author}{Lampert, C.~H.}, \bibinfo{author}{Blaschko, M.~B.}, \&
  \bibinfo{author}{Hofmann, T.} (\bibinfo{year}{2008}).
\newblock \bibinfo{title}{Beyond sliding windows: Object localization by
  efficient subwindow search}.
\newblock In {\it \bibinfo{booktitle}{Computer Vision and Pattern Recognition,
  2008. CVPR 2008. IEEE Conference on}\/} (pp. \bibinfo{pages}{1--8}).
\newblock \bibinfo{organization}{IEEE}.
\bibitem[{{Laroca} et~al.(2019){Laroca}, {Barroso}, {Diniz}, {Gon{\c{c}}alves},
  {Schwartz} \& {Menotti}}]{Laroca2019}
\bibinfo{author}{{Laroca}, R.}, \bibinfo{author}{{Barroso}, V.},
  \bibinfo{author}{{Diniz}, M.~A.}, \bibinfo{author}{{Gon{\c{c}}alves}, G.~R.},
  \bibinfo{author}{{Schwartz}, W.~R.}, \& \bibinfo{author}{{Menotti}, D.}
  (\bibinfo{year}{2019}).
\newblock \bibinfo{title}{Convolutional neural networks for automatic meter
  reading}.
\newblock {\it \bibinfo{journal}{Journal of Electronic Imaging}\/},  {\it
  \bibinfo{volume}{28}\/}, \bibinfo{pages}{1--14}.
  \DOIprefix\doi{10.1117/1.JEI.28.1.013023}.
\bibitem[{{Laroca} et~al.(2018){Laroca}, {Severo}, {Zanlorensi}, {Oliveira},
  {Gon{\c{c}}alves}, {Schwartz} \& {Menotti}}]{Laroca2018}
\bibinfo{author}{{Laroca}, R.}, \bibinfo{author}{{Severo}, E.},
  \bibinfo{author}{{Zanlorensi}, L.~A.}, \bibinfo{author}{{Oliveira}, L.~S.},
  \bibinfo{author}{{Gon{\c{c}}alves}, G.~R.}, \bibinfo{author}{{Schwartz},
  W.~R.}, \& \bibinfo{author}{{Menotti}, D.} (\bibinfo{year}{2018}).
\newblock \bibinfo{title}{A robust real-time automatic license plate
  recognition based on the {YOLO} detector}.
\newblock In {\it \bibinfo{booktitle}{International Joint Conference on Neural
  Networks (IJCNN)}\/} (pp. \bibinfo{pages}{1--10}).
\newblock \DOIprefix\doi{10.1109/IJCNN.2018.8489629}.
\bibitem[{LeCun et~al.(1998)LeCun, Bottou, Bengio \& Haffner}]{LeCun98}
\bibinfo{author}{LeCun, Y.}, \bibinfo{author}{Bottou, L.},
  \bibinfo{author}{Bengio, Y.}, \& \bibinfo{author}{Haffner, P.}
  (\bibinfo{year}{1998}).
\newblock \bibinfo{title}{Gradient-based learning applied to document
  recognition}.
\newblock {\it \bibinfo{journal}{Procs of IEEE}\/},  {\it
  \bibinfo{volume}{86}\/}, \bibinfo{pages}{2278--2324}.
\bibitem[{Lin et~al.(2017)Lin, Goyal, Girshick, He \&
  Doll{\'a}r}]{RetinaNet2017}
\bibinfo{author}{Lin, T.-Y.}, \bibinfo{author}{Goyal, P.},
  \bibinfo{author}{Girshick, R.}, \bibinfo{author}{He, K.}, \&
  \bibinfo{author}{Doll{\'a}r, P.} (\bibinfo{year}{2017}).
\newblock \bibinfo{title}{Focal loss for dense object detection}.
\newblock In {\it \bibinfo{booktitle}{Proceedings of the IEEE international
  conference on computer vision}\/} (pp. \bibinfo{pages}{2980--2988}).
\bibitem[{Lin et~al.(2014)Lin, Maire, Belongie, Hays, Perona, Ramanan,
  Doll{\'a}r \& Zitnick}]{MSCOCO}
\bibinfo{author}{Lin, T.-Y.}, \bibinfo{author}{Maire, M.},
  \bibinfo{author}{Belongie, S.}, \bibinfo{author}{Hays, J.},
  \bibinfo{author}{Perona, P.}, \bibinfo{author}{Ramanan, D.},
  \bibinfo{author}{Doll{\'a}r, P.}, \& \bibinfo{author}{Zitnick, C.~L.}
  (\bibinfo{year}{2014}).
\newblock \bibinfo{title}{Microsoft coco: Common objects in context}.
\newblock In \bibinfo{editor}{D.~Fleet}, \bibinfo{editor}{T.~Pajdla},
  \bibinfo{editor}{B.~Schiele}, \& \bibinfo{editor}{T.~Tuytelaars} (Eds.), {\it
  \bibinfo{booktitle}{Computer Vision -- ECCV 2014}\/} (pp.
  \bibinfo{pages}{740--755}).
\newblock \bibinfo{address}{Cham}: \bibinfo{publisher}{Springer International
  Publishing}.
\bibitem[{Liu et~al.(2004)Liu, Sako \& Fujisawa}]{Liu2004}
\bibinfo{author}{Liu, C.-L.}, \bibinfo{author}{Sako, H.}, \&
  \bibinfo{author}{Fujisawa, H.} (\bibinfo{year}{2004}).
\newblock \bibinfo{title}{Effects of classifier structures and training regimes
  on integrated segmentation and recognition of handwritten numeral strings}.
\newblock {\it \bibinfo{journal}{IEEE Trans. on Pattern Analysis and Machine
  Intelligence}\/},  {\it \bibinfo{volume}{26}\/}, \bibinfo{pages}{1395--1407}.
\bibitem[{Matan et~al.(1992)Matan, Burges, LeCun \& Denker}]{Matan92}
\bibinfo{author}{Matan, O.}, \bibinfo{author}{Burges, J.~C.},
  \bibinfo{author}{LeCun, Y.}, \& \bibinfo{author}{Denker, J.~S.}
  (\bibinfo{year}{1992}).
\newblock \bibinfo{title}{Multi-digit recognition using a space displacement
  neural network}.
\newblock In \bibinfo{editor}{J.~E. Moody}, \bibinfo{editor}{S.~J. Hanson}, \&
  \bibinfo{editor}{R.~L. Lippmann} (Eds.), {\it \bibinfo{booktitle}{Advances in
  Neural Information Processing Systems}\/} (pp. \bibinfo{pages}{488--495}).
\newblock \bibinfo{publisher}{Morgan Kaufmann} volume~\bibinfo{volume}{4}.
\bibitem[{Oliveira et~al.(2000)Oliveira, Lethelier, Bortolozzi \&
  Sabourin}]{Oliveira00-2}
\bibinfo{author}{Oliveira, L.~S.}, \bibinfo{author}{Lethelier, E.},
  \bibinfo{author}{Bortolozzi, F.}, \& \bibinfo{author}{Sabourin, R.}
  (\bibinfo{year}{2000}).
\newblock \bibinfo{title}{A new approach to segment handwritten digits}.
\newblock In {\it \bibinfo{booktitle}{Proc. of 7$^{th}$ International Workshop
  on Frontiers of Handwriting Recognition}\/} (pp. \bibinfo{pages}{577--582}).
\newblock \bibinfo{address}{Amsterdam, Netherlands}.
\bibitem[{Oliveira \& Sabourin(2004)}]{Oliveira2004}
\bibinfo{author}{Oliveira, L.~S.}, \& \bibinfo{author}{Sabourin, R.}
  (\bibinfo{year}{2004}).
\newblock \bibinfo{title}{Support vector machines for handwritten numerical
  string recognition}.
\newblock In {\it \bibinfo{booktitle}{9th International Workshop on Frontiers
  in Handwriting Recognition}\/} (pp. \bibinfo{pages}{39--44}).
\bibitem[{Oliveira et~al.(2002)Oliveira, Sabourin, Bortolozzi \&
  Suen}]{Oliveira02b}
\bibinfo{author}{Oliveira, L.~S.}, \bibinfo{author}{Sabourin, R.},
  \bibinfo{author}{Bortolozzi, F.}, \& \bibinfo{author}{Suen, C.~Y.}
  (\bibinfo{year}{2002}).
\newblock \bibinfo{title}{Automatic recognition of handwritten numerical
  strings: A recognition and verification strategy}.
\newblock {\it \bibinfo{journal}{IEEE Trans. on Pattern Analysis on Machine
  Intelligence}\/},  {\it \bibinfo{volume}{24}\/}, \bibinfo{pages}{1438--1454}.
\bibitem[{Pal et~al.(2003)Pal, Belaid \& Choisy}]{Pal03}
\bibinfo{author}{Pal, U.}, \bibinfo{author}{Belaid, A.}, \&
  \bibinfo{author}{Choisy, C.} (\bibinfo{year}{2003}).
\newblock \bibinfo{title}{Touching numeral segmentation using water reservoir
  concept}.
\newblock {\it \bibinfo{journal}{Pattern Recognition Letters}\/},  {\it
  \bibinfo{volume}{24}\/}, \bibinfo{pages}{261--272}.
\bibitem[{Procter et~al.(1998)Procter, Illingworth \& Elms}]{Procter98}
\bibinfo{author}{Procter, S.}, \bibinfo{author}{Illingworth, J.}, \&
  \bibinfo{author}{Elms, A.~J.} (\bibinfo{year}{1998}).
\newblock \bibinfo{title}{The recognition of handwritten digit strings of
  unknown length using hidden {M}arkov models}.
\newblock In {\it \bibinfo{booktitle}{Proc. of 14$^{th}$ International
  Conference Pattern Recognition (ICPR)}\/} (pp. \bibinfo{pages}{1515--1517}).
\bibitem[{Redmon et~al.(2016)Redmon, Divvala, Girshick \& Farhadi}]{YOLO2016}
\bibinfo{author}{Redmon, J.}, \bibinfo{author}{Divvala, S.},
  \bibinfo{author}{Girshick, R.}, \& \bibinfo{author}{Farhadi, A.}
  (\bibinfo{year}{2016}).
\newblock \bibinfo{title}{You only look once: Unified, real-time object
  detection}.
\newblock In {\it \bibinfo{booktitle}{Proceedings of the IEEE conference on
  computer vision and pattern recognition}\/} (pp. \bibinfo{pages}{779--788}).
\bibitem[{Redmon \& Farhadi(2017)}]{YOLO2017}
\bibinfo{author}{Redmon, J.}, \& \bibinfo{author}{Farhadi, A.}
  (\bibinfo{year}{2017}).
\newblock \bibinfo{title}{Yolo9000: Better, faster, stronger}.
\newblock In {\it \bibinfo{booktitle}{Computer Vision and Pattern Recognition
  (CVPR), 2017 IEEE Conference on}\/} (pp. \bibinfo{pages}{6517--6525}).
\newblock \bibinfo{organization}{IEEE}.
\bibitem[{Ren et~al.(2015)Ren, He, Girshick \& Sun}]{FasterRCNN}
\bibinfo{author}{Ren, S.}, \bibinfo{author}{He, K.}, \bibinfo{author}{Girshick,
  R.}, \& \bibinfo{author}{Sun, J.} (\bibinfo{year}{2015}).
\newblock \bibinfo{title}{Faster r-cnn: Towards real-time object detection with
  region proposal networks}.
\newblock In {\it \bibinfo{booktitle}{Proceedings of the 28th International
  Conference on Neural Information Processing Systems - Volume 1}\/} NIPS'15
  (pp. \bibinfo{pages}{91--99}).
\newblock \bibinfo{address}{Cambridge, MA, USA}: \bibinfo{publisher}{MIT
  Press}.
\bibitem[{Ribas et~al.(2013)Ribas, Oliveira, Britto \& Sabourin}]{Ribas2013}
\bibinfo{author}{Ribas, F.~C.}, \bibinfo{author}{Oliveira, L.~S.},
  \bibinfo{author}{Britto, A.~S.}, \& \bibinfo{author}{Sabourin, R.}
  (\bibinfo{year}{2013}).
\newblock \bibinfo{title}{Handwritten digit segmentation: A comparative study}.
\newblock {\it \bibinfo{journal}{International Journal on Document Analysis and
  Recognition}\/},  {\it \bibinfo{volume}{16}\/}, \bibinfo{pages}{567--578}.
\bibitem[{Roy et~al.(2016)Roy, Bhunia, Das, Dey \& Pal}]{Roy2016}
\bibinfo{author}{Roy, P.}, \bibinfo{author}{Bhunia, A.}, \bibinfo{author}{Das,
  A.}, \bibinfo{author}{Dey, P.}, \& \bibinfo{author}{Pal, U.}
  (\bibinfo{year}{2016}).
\newblock \bibinfo{title}{{HMM}-based {I}ndic handwritten word recognition
  using zone segmentation}.
\newblock {\it \bibinfo{journal}{Pattern Recognition}\/},  {\it
  \bibinfo{volume}{60}\/}, \bibinfo{pages}{1057--1075}.
\bibitem[{Russakovsky et~al.(2015)Russakovsky, Deng, Su, Krause, Satheesh, Ma,
  Huang, Karpathy, Khosla, Bernstein, Berg \& Fei-Fei}]{ILSVRC15}
\bibinfo{author}{Russakovsky, O.}, \bibinfo{author}{Deng, J.},
  \bibinfo{author}{Su, H.}, \bibinfo{author}{Krause, J.},
  \bibinfo{author}{Satheesh, S.}, \bibinfo{author}{Ma, S.},
  \bibinfo{author}{Huang, Z.}, \bibinfo{author}{Karpathy, A.},
  \bibinfo{author}{Khosla, A.}, \bibinfo{author}{Bernstein, M.},
  \bibinfo{author}{Berg, A.~C.}, \& \bibinfo{author}{Fei-Fei, L.}
  (\bibinfo{year}{2015}).
\newblock \bibinfo{title}{{ImageNet Large Scale Visual Recognition Challenge}}.
\newblock {\it \bibinfo{journal}{International Journal of Computer Vision
  (IJCV)}\/},  {\it \bibinfo{volume}{115}\/}, \bibinfo{pages}{211--252}.
  \DOIprefix\doi{10.1007/s11263-015-0816-y}.
\bibitem[{{Saabni}(2016)}]{Saabni2016}
\bibinfo{author}{{Saabni}, R.} (\bibinfo{year}{2016}).
\newblock \bibinfo{title}{Recognizing handwritten single digits and digit
  strings using deep architecture of neural networks}.
\newblock In {\it \bibinfo{booktitle}{2016 Third International Conference on
  Artificial Intelligence and Pattern Recognition (AIPR)}\/} (pp.
  \bibinfo{pages}{1--6}).
\newblock \DOIprefix\doi{10.1109/ICAIPR.2016.7585206}.
\bibitem[{Sabour et~al.(2017)Sabour, Frosst \& Hinton}]{Sabour2017}
\bibinfo{author}{Sabour, S.}, \bibinfo{author}{Frosst, N.}, \&
  \bibinfo{author}{Hinton, G.} (\bibinfo{year}{2017}).
\newblock \bibinfo{title}{Dynamic routing between capsules}.
\newblock In {\it \bibinfo{booktitle}{Advances in Neural Information Processing
  Systems 30 (NIPS 2017)}\/}.
\bibitem[{Sadri et~al.(2007)Sadri, Suen \& Bui}]{Sadri2007}
\bibinfo{author}{Sadri, J.}, \bibinfo{author}{Suen, C.~Y.}, \&
  \bibinfo{author}{Bui, T.~D.} (\bibinfo{year}{2007}).
\newblock \bibinfo{title}{A genetic framework using contextual knowledge for
  segmentation and recognition of handwritten numeral strings}.
\newblock {\it \bibinfo{journal}{Pattern Recognition}\/},  {\it
  \bibinfo{volume}{40}\/}, \bibinfo{pages}{898--919}.
\bibitem[{{Schuster} \& {Paliwal}(1997)}]{Schuster1997}
\bibinfo{author}{{Schuster}, M.}, \& \bibinfo{author}{{Paliwal}, K.~K.}
  (\bibinfo{year}{1997}).
\newblock \bibinfo{title}{Bidirectional recurrent neural networks}.
\newblock {\it \bibinfo{journal}{IEEE Transactions on Signal Processing}\/},
  {\it \bibinfo{volume}{45}\/}, \bibinfo{pages}{2673--2681}.
  \DOIprefix\doi{10.1109/78.650093}.
\bibitem[{Shi et~al.(2017)Shi, Bai \& Yao}]{Shi2017-CRNN}
\bibinfo{author}{Shi, B.}, \bibinfo{author}{Bai, X.}, \& \bibinfo{author}{Yao,
  C.} (\bibinfo{year}{2017}).
\newblock \bibinfo{title}{An end-to-end trainable neural network for
  image-based sequence recognition and its application to scene text
  recognition}.
\newblock {\it \bibinfo{journal}{IEEE Transactions on Pattern Analysis and
  Machine Intelligence}\/},  {\it \bibinfo{volume}{39}\/},
  \bibinfo{pages}{2298--2304}. \DOIprefix\doi{10.1109/TPAMI.2016.2646371}.
\bibitem[{Shi \& Govindaraju(1997)}]{Shi97}
\bibinfo{author}{Shi, Z.}, \& \bibinfo{author}{Govindaraju, V.}
  (\bibinfo{year}{1997}).
\newblock \bibinfo{title}{Segmentation and recognition of connected handwritten
  numeral strings}.
\newblock {\it \bibinfo{journal}{Pattern Recognition}\/},  {\it
  \bibinfo{volume}{30}\/}, \bibinfo{pages}{1501--1504}.
\bibitem[{Tamen et~al.(2017)Tamen, Drias \& Boughaci}]{Tamen2017}
\bibinfo{author}{Tamen, Z.}, \bibinfo{author}{Drias, H.}, \&
  \bibinfo{author}{Boughaci, D.} (\bibinfo{year}{2017}).
\newblock \bibinfo{title}{An efficient multiple classifier system for arabic
  handwritten words recognition}.
\newblock {\it \bibinfo{journal}{Pattern Recognition Letters}\/},  {\it
  \bibinfo{volume}{93}\/}.
\bibitem[{Uijlings et~al.(2013)Uijlings, van~de Sande, Gevers \&
  Smeulders}]{SelectiveSearch}
\bibinfo{author}{Uijlings, J. R.~R.}, \bibinfo{author}{van~de Sande, K. E.~A.},
  \bibinfo{author}{Gevers, T.}, \& \bibinfo{author}{Smeulders, A. W.~M.}
  (\bibinfo{year}{2013}).
\newblock \bibinfo{title}{Selective search for object recognition}.
\newblock {\it \bibinfo{journal}{International Journal of Computer Vision}\/},
  {\it \bibinfo{volume}{104}\/}, \bibinfo{pages}{154--171}. \URLprefix
  \url{https://doi.org/10.1007/s11263-013-0620-5}.
  \DOIprefix\doi{10.1007/s11263-013-0620-5}.
\bibitem[{Vellasques et~al.(2008)Vellasques, Oliveira, Britto, Koerich \&
  Sabourin}]{Vellasques2008}
\bibinfo{author}{Vellasques, E.}, \bibinfo{author}{Oliveira, L.~S.},
  \bibinfo{author}{Britto, A.~S.}, \bibinfo{author}{Koerich, A.}, \&
  \bibinfo{author}{Sabourin, R.} (\bibinfo{year}{2008}).
\newblock \bibinfo{title}{Filtering segmentation cuts for digit string
  recognition}.
\newblock {\it \bibinfo{journal}{Pattern Recognition}\/},  {\it
  \bibinfo{volume}{41}\/}, \bibinfo{pages}{3044--3053}.
\bibitem[{Voigtlaender et~al.(2016)Voigtlaender, Doetsch \&
  Ney}]{Voigtlaender2016}
\bibinfo{author}{Voigtlaender, P.}, \bibinfo{author}{Doetsch, P.}, \&
  \bibinfo{author}{Ney, H.} (\bibinfo{year}{2016}).
\newblock \bibinfo{title}{Handwriting recognition with large multidimensional
  long short-term memory recurrent neural networks}.
\newblock In {\it \bibinfo{booktitle}{2016 15th International Conference on
  Frontiers in Handwriting Recognition (ICFHR)}\/} (pp.
  \bibinfo{pages}{228--233}).
\newblock \DOIprefix\doi{10.1109/ICFHR.2016.0052}.
\bibitem[{Wang et~al.(2000)Wang, Govindaraju \& Srihari}]{Wang00}
\bibinfo{author}{Wang, X.}, \bibinfo{author}{Govindaraju, V.}, \&
  \bibinfo{author}{Srihari, S.~N.} (\bibinfo{year}{2000}).
\newblock \bibinfo{title}{Holistic recognition of handwritten character pairs}.
\newblock {\it \bibinfo{journal}{Pattern Recognition}\/},  {\it
  \bibinfo{volume}{33}\/}, \bibinfo{pages}{1967--1973}.
\bibitem[{Xiao et~al.(2017)Xiao, Jin, Y.Yang, Yang, Sun \& Chang}]{Xiao2017}
\bibinfo{author}{Xiao, X.}, \bibinfo{author}{Jin, L.},
  \bibinfo{author}{Y.Yang}, \bibinfo{author}{Yang, W.}, \bibinfo{author}{Sun,
  J.}, \& \bibinfo{author}{Chang, T.} (\bibinfo{year}{2017}).
\newblock \bibinfo{title}{Building fast and compact convolutional neural
  networks for offline handwritten {C}hinese character recognition}.
\newblock {\it \bibinfo{journal}{Pattern Recognition}\/},  {\it
  \bibinfo{volume}{72-81}\/}.
\bibitem[{{Xu} et~al.(2018){Xu}, {Zhou} \& {Zhang}}]{Xu2018}
\bibinfo{author}{{Xu}, X.}, \bibinfo{author}{{Zhou}, J.}, \&
  \bibinfo{author}{{Zhang}, H.} (\bibinfo{year}{2018}).
\newblock \bibinfo{title}{Screen-rendered text images recognition using a deep
  residual network based segmentation-free method}.
\newblock In {\it \bibinfo{booktitle}{2018 24th International Conference on
  Pattern Recognition (ICPR)}\/} (pp. \bibinfo{pages}{2741--2746}).
\newblock \DOIprefix\doi{10.1109/ICPR.2018.8545678}.
\bibitem[{Y.~Wua \& Liu(2017)}]{Wua2017}
\bibinfo{author}{Y.~Wua, F.~Y.}, \& \bibinfo{author}{Liu, C.~L.}
  (\bibinfo{year}{2017}).
\newblock \bibinfo{title}{Improving handwritten {C}hinese text recognition
  using neural network language models and convolutional neural network shape
  models}.
\newblock {\it \bibinfo{journal}{Pattern Recognition}\/},  {\it
  \bibinfo{volume}{65}\/}, \bibinfo{pages}{251--264}.
\bibitem[{Zhan et~al.(2017)Zhan, Wang \& Lu}]{Zhan2017}
\bibinfo{author}{Zhan, H.}, \bibinfo{author}{Wang, Q.}, \& \bibinfo{author}{Lu,
  Y.} (\bibinfo{year}{2017}).
\newblock \bibinfo{title}{Handwritten digit string recognition by combination
  of residual network and rnn-ctc}.
\newblock In \bibinfo{editor}{D.~Liu}, \bibinfo{editor}{S.~Xie},
  \bibinfo{editor}{Y.~Li}, \bibinfo{editor}{D.~Zhao}, \&
  \bibinfo{editor}{E.-S.~M. El-Alfy} (Eds.), {\it \bibinfo{booktitle}{Neural
  Information Processing}\/} (pp. \bibinfo{pages}{583--591}).
\newblock \bibinfo{address}{Cham}: \bibinfo{publisher}{Springer International
  Publishing}.
\bibitem[{Ziyong et~al.(2017)Ziyong, Zhaoyang, Shuanping \& Jun}]{Ziyong2017}
\bibinfo{author}{Ziyong, F.}, \bibinfo{author}{Zhaoyang, Y.},
  \bibinfo{author}{Shuanping, J. L.~H.}, \& \bibinfo{author}{Jun, S.}
  (\bibinfo{year}{2017}).
\newblock \bibinfo{title}{Robust shared feature learning for script and
  handwritten/machine-printed identification}.
\newblock {\it \bibinfo{journal}{Pattern Recognition Letters}\/},  {\it
  \bibinfo{volume}{100}\/}.

\end{thebibliography}
\end{document}